\documentclass[nohyperref]{article}
\usepackage{hyperref}
\usepackage[accepted]{icml2023}
\usepackage{microtype}
\usepackage{graphicx}
\usepackage{subfigure}
\usepackage{booktabs} % for professional tables
\usepackage{multicol,multirow}
\usepackage{bbding}
\usepackage{amsmath}
\usepackage{amssymb}
\usepackage{mathtools}
\usepackage{amsthm}
\usepackage{tcolorbox}
\usepackage{wrapfig}
\definecolor{dkblue}{rgb}{0,0.08,0.45}

\newcommand{\blue}[1]{\textcolor{dkblue}{#1}}

\renewcommand{\algorithmicrequire}{\textbf{Input:}}
\renewcommand{\algorithmicensure}{\textbf{Output:}}

\theoremstyle{plain}
\newtheorem{theorem}{Theorem}[section]
\newtheorem{proposition}[theorem]{Proposition}

\newtheorem{corollary}[theorem]{Corollary}
\theoremstyle{definition}

\newtheorem{assumption}[theorem]{Condition}
\theoremstyle{remark}

\DeclareMathOperator*{\argmin}{arg\,min}
\DeclareMathOperator*{\argmax}{arg\,max}
\newcommand{\meanN}{\frac{1}{n} \sum_{i=1}^n}

\newcommand{\sumA}{\sum_{a \in \mathcal{A}}}

\newcommand{\sumC}{\sum_{c \in \mathcal{C}}}
\newcommand{\mE}{\mathbb{E}}
\newcommand{\mV}{\mathbb{V}}
\newcommand{\mC}{\mathrm{Cov}}

\newcommand{\calD}{\mathcal{D}}
\newcommand{\calX}{\mathcal{X}}
\newcommand{\calA}{\mathcal{A}}

\newcommand{\calC}{\mathcal{C}}
\newcommand{\calE}{\mathcal{E}}

\newcommand{\nablat}{\nabla_{\theta}}
\newcommand{\score}{s_{\theta}}
\newcommand{\firstpolicy}{\pi_{\theta}^{1st}}
\newcommand{\secondpolicy}{\pi_{\psi}^{2nd}}
\newcommand{\overallpolicy}{\pi_{\theta,\psi}^{overall}}

\newcommand{\trueV}{V(\pi)}
\newcommand{\ips}{\hat{V}_{\mathrm{IPS}} (\pi; \calD)}
\newcommand{\mips}{\hat{V}_{\mathrm{MIPS}} (\pi; \calD)}
\newcommand{\dr}{\hat{V}_{\mathrm{DR}} (\pi; \calD, \hat{q})}
\newcommand{\dm}{\hat{V}_{\mathrm{DM}} (\pi; \calD, \hat{q})}
\newcommand{\offcem}{\hat{V}_{\mathrm{OffCEM}} (\pi; \calD)}

\newcommand{\truePG}{\nabla_{\theta}V(\pi_{\theta})}
\newcommand{\ipsPG}{\nabla_{\theta} \widehat{V}_{\mathrm{IPS}} (\pi_{\theta}; \calD)}
\newcommand{\drPG}{\nabla_{\theta} \widehat{V}_{\mathrm{DR}} (\pi_{\theta}; \calD)}
\newcommand{\sipsPG}{\nabla_{\theta} \widehat{V}_{\mathrm{sIPS}} (\pi_{\theta}; \calD)}
\newcommand{\potec}{\nabla_{\theta} \widehat{V}_{\mathrm{POTEC}} (\overallpolicy; \calD)}
\newcommand{\potecone}{\nabla_{\theta} \widehat{V}_{\mathrm{POTEC1}} (\pi_{\theta}; \calD)}

\newcommand{\biasPOTEC}{\mathrm{Bias} ( \potec )}

\icmltitlerunning{POTEC: Off-Policy Learning for Large Action Spaces via Two-Stage Policy Decomposition}

\begin{document}

\twocolumn[
\icmltitle{POTEC: Off-Policy Learning for Large Action Spaces\\ via Two-Stage Policy Decomposition}

\begin{icmlauthorlist}
\icmlauthor{Yuta Saito}{cornell}
\icmlauthor{Jihan Yao}{was}
\icmlauthor{Thorsten Joachims}{cornell}
\end{icmlauthorlist}
\icmlaffiliation{cornell}{Department of Computer Science, Cornell University, NY, USA}
\icmlaffiliation{was}{Paul G. Allen School of Computer Science \& Engineering, University of Washington, Seattle, WA, USA}
\icmlcorrespondingauthor{Yuta Saito}{ys552@cornell.edu}
\icmlcorrespondingauthor{Thorsten Joachims}{tj@cs.cornell.edu}

\vskip 0.3in
]

\printAffiliationsAndNotice{} 

\begin{abstract}
We study off-policy learning (OPL) of contextual bandit policies in large discrete action spaces where existing methods -- most of which rely crucially on reward-regression models or importance-weighted policy gradients -- fail due to excessive bias or variance. To overcome these issues in OPL, we propose a novel \textit{two-stage} algorithm, called \textit{\textbf{P}olicy \textbf{O}ptimization via \textbf{T}wo-Stage Policy D\textbf{ec}omposition (POTEC)}. It leverages clustering in the action space and learns two different policies via policy- and regression-based approaches, respectively. In particular, we derive a novel low-variance gradient estimator that enables to learn a first-stage policy for cluster selection efficiently via a policy-based approach. To select a specific action within the cluster sampled by the first-stage policy, POTEC uses a second-stage policy derived from a regression-based approach within each cluster. We show that a local correctness condition, which only requires that the regression model preserves the relative expected reward differences of the actions within each cluster, ensures that our policy-gradient estimator is unbiased and the second-stage policy is optimal. We also show that POTEC provides a strict generalization of policy- and regression-based approaches and their associated assumptions. Comprehensive experiments demonstrate that POTEC provides substantial improvements in OPL effectiveness particularly in large and structured action spaces. 
\end{abstract}

\section{Introduction}
Many interactive systems (e.g., voice assistants, ad-placement, recommender systems) are increasingly controlled by algorithms that learn from historical user interactions. These interactions consist of the context (e.g., user profile, query), the action chosen by the logging policy (e.g., recommended product), and the resulting reward (e.g., click, conversion). Using such logged interactions, a common goal is to train a new policy that improves the expected reward. This \textit{off-policy learning} (OPL) task is of great practical relevance, as it enables us to improve system effectiveness without the risky, slow, and potentially unethical use of online exploration.

A highly effective approach to OPL is policy learning by estimating the policy gradient, which has resulted in a number of practical OPL methods for small action spaces~\citep{joachims2018deep,metelli2021subgaussian,su2020doubly,su2019cab,swaminathan2015batch,swaminathan2015counterfactual}. Unfortunately, this policy-based approach can deteriorate dramatically for large action spaces, which are prevalent in many potential applications of OPL where there exist millions of items (e.g., recommendations of movies, songs, products). In particular, in such large-scale environments, existing policy-based methods, which are mostly based on importance-weighted policy gradients, can collapse due to extremely large variance~\citep{saito2022off,saito2023off}. While regression-based approaches to OPL, which learn the expected reward function and choose the action with the highest predicted reward, could potentially circumvent the variance issue, they are known to suffer from high bias due to model misspecification~\citep{farajtabar2018more,sachdeva2020off,voloshin2019empirical,saito2021open} and thus do not provide a readily available solution either.

To overcome this bias and variance dilemma of OPL arising particularly in large action spaces, we develop a novel \textit{two-stage} OPL algorithm called \textit{\textbf{P}olicy \textbf{O}ptimization via \textbf{T}wo-Stage Policy D\textbf{ec}omposition (\textbf{POTEC})}. POTEC operates under a novel policy decomposition framework, wherein the typical overall policy (marginal action distribution) is decomposed into first-stage and second-stage policies via an action cluster space. The first-stage policy focuses on identifying promising action clusters (cluster distribution), while the second-stage policy aims to select the optimal action within a specific cluster sampled from the first-stage policy (conditional action distribution). A key feature of POTEC is its distinct learning approaches for these policies. The first-stage policy is learned using a policy-based approach with a novel policy gradient estimator, called the POTEC gradient estimator. The POTEC gradient estimator combines importance weighting in the action cluster space to estimate the value of clusters while using a \textit{pairwise} reward model to deal with the effect of individual actions within each cluster. We show that our gradient estimator is unbiased under \textit{local correctness}~\cite{saito2023off}, requiring only that the regression model accurately preserves the relative reward differences within each action cluster. We also show that we can be based on the same reward regression model used in the POTEC gradient estimator to readily construct a second-stage policy through a regression-based approach.

Compared to standard policy-based methods, the POTEC gradient estimator for the first-stage policy exhibits significantly lower variance in large action spaces, as it applies importance weighting to only the action cluster space, which is considerably more compact than the original action space. Furthermore, POTEC is expected to be more resilient to estimation bias than typical regression-based approaches, since our first-stage policy is based on an unbiased policy gradient and the second-stage policy only needs to learn the relative value differences between actions, which is less demanding than conventional absolute reward regression. Moreover, we show that POTEC and local correctness provide a full spectrum of OPL approaches whose endpoints are policy- and regression-based methods and their associated reward-modeling conditions. Experiments on synthetic and extreme classification data demonstrate that POTEC can provide substantially more effective OPL than conventional methods particularly in large and structured action spaces.

\section{Off-Policy Learning for Contextual Bandits} \label{sec:formulation}
We formulate OPL under the general contextual bandit process, where a decision maker repeatedly observes a context $x \in \calX$ drawn i.i.d.\ from an unknown distribution $p(x)$. Given context $x$, a potentially stochastic \textit{policy} $\pi(a\,|\,x)$ chooses action $a$ from a finite action space denoted as $\calA$. The reward $r \in [0, r_{\mathrm{max}}]$ is then sampled from some unknown conditional distribution $p(r\,|\,x,a)$. 
We define the \textit{value} of policy $\pi$ as a measure of its effectiveness:
\begin{align*}
    \trueV := \mE_{p(x)\pi(a|x)p(r|x,a)} [r] = \mE_{p(x)\pi(a|x)} [q (x,a)]  ,
\end{align*}
where we use $q(x,a) := \mE [r\,|\,x,a]$ to denote the reward function (the expected reward given $x$ and $a$).

Our goal is to learn a new policy $\pi_{\theta}$ parameterized by $\theta$ to maximize the policy value as $$\theta^*= \argmax_{\theta \in \Theta} V(\pi_{\theta}).$$ The logged data we can use for performing OPL takes the form $\calD := \{(x_i,a_i,r_i)\}_{i=1}^n$, which contains $n$ independent observations drawn from the logging policy $\pi_0$.

Below, we describe two typical approaches to OPL, namely the policy-based and regression-based approaches, and summarize their limitations, particularly in large action spaces.

\begin{figure*}[t]
\centering
\includegraphics[clip, width=13.5cm]{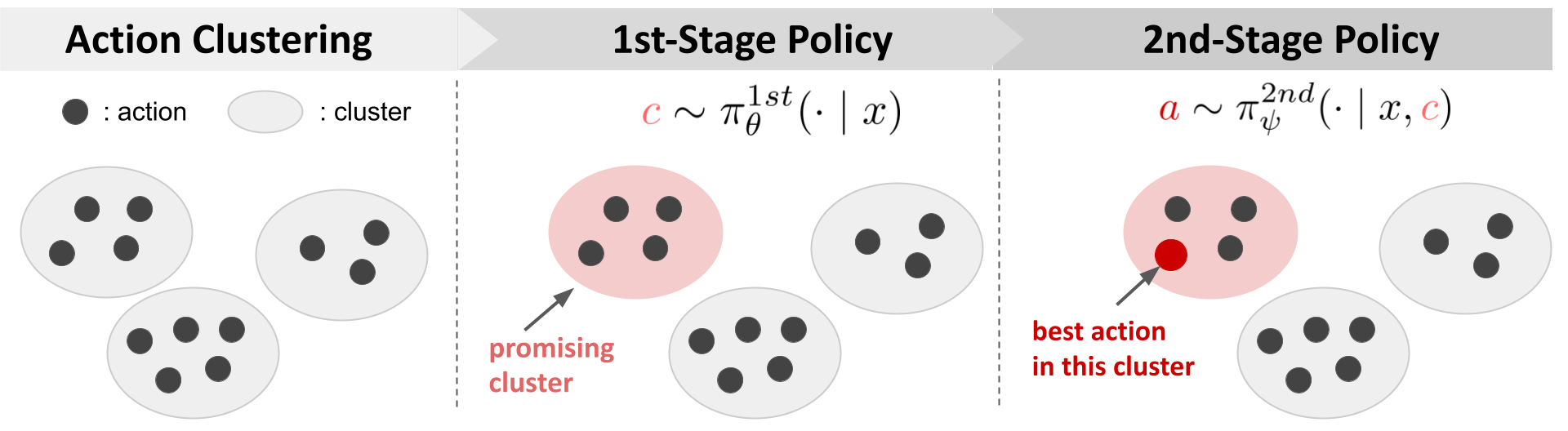}
\vspace*{-3mm}
\caption{The Two-Stage Off-Policy Learning Procedure of Our POTEC Algorithm, which first forms action clustering $c_a$, and then identifies a promising cluster by the 1st-stage policy $\firstpolicy$, and finally picks the best action in the cluster by the 2nd-stage policy $\secondpolicy$.} \label{fig:potec}
\raggedright
\end{figure*}

\textbf{The policy-based approach} learns the policy parameter via iterative gradient ascent as $\theta_{t+1} \leftarrow \theta_t + \nablat V(\pi_{\theta}) $. Since we do not know the true gradient $$\nablat V (\pi_{\theta}) = \mE_{p(x)\pi_{\theta}(a|x)}[q(x,a)\nablat\log\pi_{\theta}(a\,|\,x)],$$ we need to estimate it from the logged data. A common way to do so is to apply importance weighting as
\begin{align}
   \ipsPG 
   &:=\meanN  w(x_i,a_i) r_i \score (x_i,a_i), \label{eq:is-pg}
\end{align}
where $w(x,a):= \pi_{\theta}(a\,|\,x) / \pi_{0}(a\,|\,x)$ is the (vanilla) importance weight and $\score (x,a) := \nablat\log \pi_{\theta} (a\,|\,x) $ is the policy score function.

Eq.~\eqref{eq:is-pg} is unbiased (i.e., $\mE[\ipsPG]=\truePG$) under the following condition. 

\begin{assumption} (Full Support) \label{assumption:full_support}
The logging policy $\pi_0$ is said to have full support if $\pi_0(a\,|\,x) > 0,\; \forall (x,a) \in \calX \times \calA$.
\end{assumption}

For large action spaces, unfortunately, this requirement of full support is problematic for two reasons. First, violating the requirement can introduce substantial bias~\cite{felicioni2022off,sachdeva2020off}. Second, fulfilling the requirement for large action spaces leads to excessive variance, since $\pi_0(a\,|\,x)$ becomes small. At first glance, \textit{doubly-robust} (DR) estimation~\cite{dudik2014doubly} may appear helpful for dealing with the variance issue. 
\begin{align}
   \drPG \!:=\! \frac{1}{n}\!\sum_{i=1}^n & w(x_i,a_i) (r_i \!-\! \hat{q} (x_i,a_i))  \score (x_i,a_i)  \notag \\
   &\!\!\!+ \!\mE_{\pi_{\theta}(a|x_i)} [ \hat{q} (x_i,a) \score (x_i,a) ]    \label{eq:dr-pg}
\end{align}
DR uses a reward function estimator $\hat{q}(x,a)$ while maintaining unbiasedness under Condition~\ref{assumption:full_support}, and its variance is often lower than that of Eq.~\eqref{eq:is-pg}. However, unless the rewards are close to deterministic and the reward estimates $\hat{q}(x,a)$ are close to perfect, its variance can still be extremely large due to vanilla importance weighting, which leads to inefficient OPL in large action spaces~\citep{saito2022off,peng2023offline,sachdeva2023off}. The issue of the IPS and DR policy gradients can be seen by calculating their variance (for a particular parameter $\theta\in\mathbb{R}^d$) as
\begin{align}
    & n \operatorname{tr} \left(\mC_{\calD} \big[  \drPG \big] \right) \notag \\
    & = \sum_{j=1}^d \Big\{ \mE_{p(x)\pi_0(a|x)} [ (w(x,a) \score^{(j)}(x,a))^2 \sigma^2 (x,a) ]  \notag \\
    & \quad + \mE_{p(x)} \left[  \mV_{\pi_0(a|x)} [ w(x,a) \Delta_{q,\hat{q}} (x,a) \score^{(j)}(x,a) ]  \right] \notag\\ 
    & \quad + \mV_{p(x)} \left[  \mE_{\pi(a|x)} [ q (x,a) \score^{(j)}(x,a) ] \right] \Big\}, \label{eq:dr_variance}
\end{align}
where $ \sigma^2 (x,a) := \mV [r\,|\,x,a] $ and $\Delta_{q,\hat{q}} (x,a) := q(x,a)-\hat{q}(x,a)$. $\score^{(j)}(x,a)$ is the $j$-th dimension of the score function. Note that the variance of IPS can be obtained by setting $\hat{q}(x,a) = 0$. The variance reduction of DR comes from the second term where $\Delta_{q,\hat{q}} (x,a)$ is smaller than $q(x,a)$ if $\hat{q}(x,a)$ is accurate. However, we can also see that the variance contributed by the first term can be extremely large for both IPS and DR when the reward is noisy and the weights $w(x,a)$ become large, which occurs when $\pi_{\theta}$ assigns large probabilities to actions that are less likely under $\pi_0$.

\textbf{The regression-based approach} employs an off-the-shelf supervised machine learning method to estimate the reward function, for example, by solving $$\theta = \argmin_{\theta} \sum_{(x,a,r) \in \calD} \ell \big(r, \hat{q}_{\theta} (x,a) \big).$$ Then, it transforms the estimated reward function $\hat{q}_{\theta} (x,a)$ into a decision-making rule, for example, by applying the softmax function $\pi_{\theta}(a\,|\,x) = \frac{\exp(\hat{q}_{\theta} (x,a) / \tau)}{\sum_{a'\in\calA}\exp(\hat{q}_{\theta} (x,a') / \tau)},$ where $\tau >0$ is a temperature parameter.  

This approach avoids the use of importance weighting and is therefore relatively robust to high variance compared to the policy-based approach, even in large action spaces. However, it is widely acknowledged that this approach may fail significantly due to bias issues resulting from the difficulity in accurately estimating the expected reward for every action in $\calA$~\citep{farajtabar2018more,voloshin2019empirical}.

\section{The POTEC Algorithm}  \label{sec:method}
The following proposes a new OPL algorithm, named \textbf{POTEC}, that circumvents the challenges of policy-based and regression-based approaches for large action spaces. As depicted in Figure~\ref{fig:potec}, POTEC leverages the following novel decomposition of an overall policy $\pi(a\,|\,x)$.
\begin{tcolorbox}
\textbf{The Two-Stage Policy Decomposition:}
\begin{align}
    \overallpolicy (a \,|\, x) = \sum_{c \in \calC} \firstpolicy (c \,|\, x) \secondpolicy (a \,|\, x,c),
    \label{eq:policy-decomposition}
\end{align}
where the marginal action-selection (overall) policy ($\overallpolicy$) is decomposed into the cluster-selection (first-stage) policy ($\firstpolicy$) and conditional action-selection (second-stage) policy ($\secondpolicy$), parametrized by $\theta$ and $\psi$ respectively. 
\end{tcolorbox}
This policy decomposition is defined via some pre-defined clustering structure in the action space, where $c_a \in \calC$ represents the cluster to which action $a$ belongs (typically $|\calC| \ll |\calA|$). There are many real-world situations where we can leverage such structured action spaces when performing OPL. For example, in a movie recommendation problem, the cluster space could capture the relevance of each genre to users. Although we consider context-independent and deterministic action clusters for brevity in the main text, our framework can easily be extended to more general types of action clustering (i.e., context-dependent and stochastic), as demonstrated in the appendix.

Leveraging this decomposition, POTEC \textbf{(i)} trains the 1st-stage policy $\firstpolicy$, a parameterized distribution over the cluster space $\calC$, via a policy-based approach, and then \textbf{(ii)} trains the 2nd-stage policy $\secondpolicy$, a parameterized distribution over the action space $\calA$ conditional on a cluster sampled by the 1st-stage policy, using a regression-based approach.

The underlying intuition is that we should be able to apply a policy-based approach to identify promising action clusters with low bias and variance since the cluster space is much smaller than the original action space. We can then apply a regression-based 2nd-stage policy to identify the promising actions within a cluster with low variance. The resulting overall policy should be more robust to reward modeling errors than the typical regression-based approach because we only apply a regression-based policy within each cluster.

When performing inference for an incoming context $x$, we first sample a cluster from the 1st-stage policy as $c \sim \firstpolicy(\cdot \,|\, x) $. We then apply the 2nd-stage policy to choose the action given the cluster as $a \sim \secondpolicy (\cdot \,|\, x,c)$. This procedure is equivalent to sampling an action from the overall policy $a \sim \overallpolicy(\cdot \,|\, x)$ induced by $\firstpolicy$ and $\secondpolicy$.

Below, we describe how to train 1st- and 2nd-stage policies to directly improve the value of the overall policy, i.e.,
\begin{align*}
   (\theta^*, \psi^*) = \argmax_{\theta,\psi}\; V(\overallpolicy).
\end{align*}

\subsection{Training the 1st-Stage Policy $\firstpolicy$} \label{sec:1st-stage-policy-training}
First, we develop a training procedure for the 1st-stage policy given a (pre-trained) 2nd-stage policy. Then, the theoretical analysis of the proposed training procedure will naturally tell us how we should construct the 2nd-stage policy (which will be described in the next subsection).

As mentioned earlier, given a (pre-trained) 2nd-stage policy $\secondpolicy$, we consider training the 1st-stage policy $\firstpolicy$, parameterized by $\theta$, via a policy-based approach as below.
\begin{align}
    \theta_{t+1} \leftarrow \theta_t + \nablat V(\overallpolicy) \label{eq:first-stage-update}
\end{align}
This performs gradient ascent of $\theta$ with the aim of improving the value of the overall policy $\overallpolicy$. The true policy gradient in Eq.~\eqref{eq:first-stage-update} is given as follows (derived in Appendix~\ref{app:proof}),
\begin{align}
    \nablat V(\overallpolicy)
    = \mE_{p(x)\firstpolicy(c|x)} \!\left[ q^{\secondpolicy}\!(x,c) \score(x,c) \right], \label{eq:true-1st-stage-pg}
\end{align}
where we use $q^{\secondpolicy}(x,c) :=  \mE_{\secondpolicy(a|x,c)} [q(x,a)]$ to denote the value of cluster $c$ under a 2nd-stage policy\footnote{This implies that the optimal cluster that should be chosen by the 1st-stage policy can be different given different 2nd-stage policies. Appendix~\ref{sec:true-1st-stage-pg} elaborates on this via a numerical example.}
and $\score(x,c) := \nablat\log \firstpolicy (c\,|\,x)$ to denote the policy score function of the 1st-stage policy. 

Hence, given a 2nd-stage policy, our objective is to estimate the policy gradient in Eq.\eqref{eq:true-1st-stage-pg} to train a 1st-stage policy. We achieve this via the following \textbf{POTEC gradient estimator},
\begin{align}
   & \potec \label{eq:potec-pg} \\
   & := \meanN \bigg\{ w(x_i,c_{a_i}) (r_i - \hat{f} (x_i,a_i) ) \score(x_i,c_{a_i}) \notag \\
   & \qquad\qquad\qquad + \mE_{\firstpolicy(c|x_i)} [ \hat{f}^{\secondpolicy} (x_i,c) \score(x_i,c) ] \bigg\} \notag ,
\end{align}
where $w(x,c):=\firstpolicy(c\,|\,x)/\pi_0(c\,|\,x) $ is the \textit{cluster importance weight} and $\hat{f}^{\secondpolicy}(x,c) :=  \mE_{\secondpolicy(a|x,c)} [\hat{f}(x,a)]$ for some given regression model $\hat{f}(x,a)$. The first term of Eq.~\eqref{eq:potec-pg} estimates the value of cluster $c$ via cluster importance weighting and the second term deals with the value of individual actions via the regression model $\hat{f}$. Since our policy gradient estimator applies importance weighting with respect to only the action cluster space, it is expected to provide a substantial reduction in variance compared to typical policy gradient estimators such as IPS and DR. Note that we will discuss how we should optimize the regression model $\hat{f}$ based on the following analysis of our gradient estimator.

First, we characterize the bias of the POTEC gradient estimator under the following full cluster support condition (which is less restrictive than Condition~\ref{assumption:full_support}).

\begin{assumption} (Full Cluster Support) \label{assumption:full_cluster_support}
The logging policy $\pi_0$ has full cluster support if $\pi_0(c\,|\,x) > 0, \; \forall (x,c) \in \calX\times\calC$.
\end{assumption}

In the following theorem, we denote with $\Delta_q(x,a,b) := q(x,a) - q(x,b)$ the difference in the expected rewards between the pair of actions $a$ and $b$ given $x$, which we call the \textbf{relative value difference} of the actions. $\Delta_{\hat{f}} (x,a,b):= \hat{f}(x,a) - \hat{f}(x,b)$ is an estimate of the relative value difference between $a$ and $b$ based on $\hat{f}$.

\begin{theorem} (Bias Analysis) \label{thm:bias}
If Condition~\ref{assumption:full_cluster_support} is true, the POTEC gradient estimator in Eq.~\eqref{eq:potec-pg} has the following bias for some given regression model $\hat{f}(x,a)$,
\begin{align}
    & \biasPOTEC \label{eq:bias-of-potec} \\
    & = \mE_{p(x)\pi_0^{1st}(c|x)} \Big[\sum_{a<b:c_a=c_b=c} \pi_0^{2nd}(a\,|\,x,c) \pi_0^{2nd}(b\,|\,x,c) \notag \\
    & \big( \Delta_{q} (x,a,b) - \Delta_{\hat{f}} (x,a,b) \big)  \left( w(x,b) - w(x,a) \right) \score(x,c) \Big],\notag
\end{align}
where $a,b \in \calA$. 
\end{theorem}

The proof is given in Appendix~\ref{app:bias}. Theorem~\ref{thm:bias} shows that the bias of the POTEC gradient estimator is characterized by the \textit{accuracy of the regression model $\hat{f}$ with respect to the relative value difference}, which is quantified by $\Delta_q(x,a,b) - \Delta_{\hat{f}}(x,a,b)$. When $\hat{f}$ preserves the relative value difference of the actions within each cluster accurately, the second factor in Eq.~\eqref{eq:bias-of-potec} becomes small and so does the bias of the POTEC gradient estimator. This also suggests that, in an ideal case when the following local correctness condition~\cite{saito2023off} is satisfied, the POTEC gradient estimator becomes unbiased.

\begin{assumption} (Local Correctness) \label{assumption:local_correctness}
A regression model and action clustering satisfy local correctness if $\Delta_q(x,a,b) = \Delta_{\hat{f}} (x,a,b)$ for all $x\in\calX$ and $a,b\in\calA$ s.t. $c_a=c_b$.
\end{assumption}

\begin{corollary} (Unbiasedness of POTEC) \label{cor:unbiased}
Under Conditions~\ref{assumption:full_cluster_support} and~\ref{assumption:local_correctness}, the POTEC gradient estimator is unbiased for the true policy gradient in Eq.~\eqref{eq:true-1st-stage-pg}, i.e., $\mE_{\mathcal{D}}[\potec] = \nablat V(\overallpolicy)$.
\end{corollary}

The above analysis implies that, in terms of bias minimization, we should optimize the regression model in a way that preserves the relative value difference of the actions within each cluster, i.e., small $|\Delta_{q} (x,a,b) - \Delta_{\hat{f}} (x,a,b)|$. 

Next, the following shows the variance of the POTEC gradient estimator, which tells us how we should optimize the regression model $\hat{f}$ regarding variance minimization.

\begin{proposition} (Variance Analysis) \label{prop:variance}
Under Conditions~\ref{assumption:full_cluster_support} and~\ref{assumption:local_correctness}, for a particular parameter $\theta \in \mathbb{R}^d$, the POTEC gradient estimator has the following variance.
\begin{align}
    & n \operatorname{tr} \left(\mC_{\calD} \big[  \potec \big]\right)  \notag \\
    & = \sum_{j=1}^d \Big\{ \mE_{p(x)\pi_0(a|x)} \left[ (w(x,c_a) \score^{(j)}(x,c_a))^2 \sigma^2(x,a) \right] \notag \\
    & \quad + \mE_{p(x)} \left[ \mV_{\pi_0(a|x)} \left[  w(x,c_a) \Delta_{q,\hat{f}}(x,a) \score^{(j)}(x,c_a) \right] \right] \notag \\
    & \quad + \mV_{p(x)} \left[ \mE_{\firstpolicy(c|x)} \left[ q^{\secondpolicy}(x,c) \score^{(j)}(x,c) \right] \right] \Big\}, \label{eq:POTEC_variance}
\end{align}
where $\Delta_{q,\hat{f}}(x,a):= q(x,a) - \hat{f}(x,a)$ is the error of $\hat{f}(x,a)$ against $q(x,a)$. See Appendix~\ref{app:variance} for the proof.
\end{proposition}

Proposition~\ref{prop:variance} shows that the variance of the POTEC gradient estimator depends only on $w(x,c)$ rather than $w(x,a)$, implying reduced variance compared to IPS and DR (c.f., Eq.~\eqref{eq:dr_variance}). It also suggests that, in terms of variance minimization, we should optimize the regression model in a way that minimizes $|\Delta_{q,\hat{f}}(x,a)|$ compared to minimizing $|\Delta_{q} (x,a,b) - \Delta_{\hat{f}} (x,a,b)|$ for the bias. 

Therefore, in order to optimize the statistical properties of the POTEC gradient estimator, we should ideally optimize the regression model via the following two-step procedure.
\paragraph{1. Bias Minimization Step:}
Optimize the pairwise regression function $\hat{h}_{\psi}: \calX \times \calA \rightarrow \mathbb{R}$, parameterized by $\psi$, to approximate the relative value difference $\Delta_q(x,a,b)$ via
        \begin{align}
            \min_{\psi} \!\!\!\!\! \sum_{(x,a,b,r_a,r_b) \in \calD_{pair}} \!\!\!\!\!\!\!\!\!\! \ell_h \left(r_{a} - r_{b}, \hat{h}_{\psi} (x,a) -  \hat{h}_{\psi} (x,b) \!\right), \label{eq:bias-minimization}
        \end{align}
    where $\calD_{pair}$ is a dataset augmented for performing pairwise regression, which is defined as
    $$ \calD_{pair} := \Big\{ (x,a,b,r_a,r_b)  \,|\, \begin{array}{ll} (x_a,a,r_a), (x_b,b,r_b) \in \calD  \\ \quad x=x_a=x_b, c_a = c_b \end{array} \Big\}.$$
\paragraph{2. Variance Minimization Step:} 
Optimize the baseline function $\hat{g}_{\omega}: \calX \times \calC \rightarrow \mathbb{R}$, parameterized by $\omega$, to minimize $\Delta_{q,\hat{f}}(x,a)$ given $\hat{f}(x,a) = \hat{g}_{\omega}(x,c_a) + \hat{h}_{\psi}(x,a)$ via
    \begin{align}
        \min_{\omega} \sum_{(x,a,r) \in \calD} \ell_g \left(r - \hat{h}_{\psi} (x,a), \hat{g}_{\omega} (x,c_a) \right). \label{eq:variance-minimization}
    \end{align}
 $\ell_h, \ell_g: \mathbb{R} \times \mathbb{R} \rightarrow \mathbb{R}_{\ge 0}$ are some appropriate loss functions such as squared loss. As suggested in our analysis, $\hat{h}_{\psi}(x,a)$ fully characterizes the bias of the POTEC gradient estimator, and thus the second step can fully commit to variance minimization by optimizing the baseline function $\hat{g}_{\omega}(x,c_a)$, which does not affect the bias of the POTEC gradient estimator. We can then construct our regression model as $\hat{f}_{\psi,\omega}(x,a) = \hat{g}_{\omega}(x,c_a) + \hat{h}_{\psi}(x,a)$. Even if the two-step procedure is infeasible due to insufficient pairwise data, we can still perform a conventional regression for the expected absolute reward to directly optimize the parameterized function (globally or separately for each cluster) $\hat{f}_{\omega}: \calX \times \calA \rightarrow \mathbb{R}$ via $\min_{\omega} \sum_{(x,a,r) \in \calD} \ell_f \big(r, \hat{f}_{\omega} (x,a) \big)$ and then use $\hat{f}_{\omega}$ in Eq.~\eqref{eq:potec-pg}. Even for such a conventionally trained regression model $\hat{f}_{\omega}$, the POTEC estimator still has advantages over existing policy gradient estimators, such as IPS and DR, due to its substantially reduced variance.

 \begin{figure*}[t]
\centering
\includegraphics[clip, width=13cm]{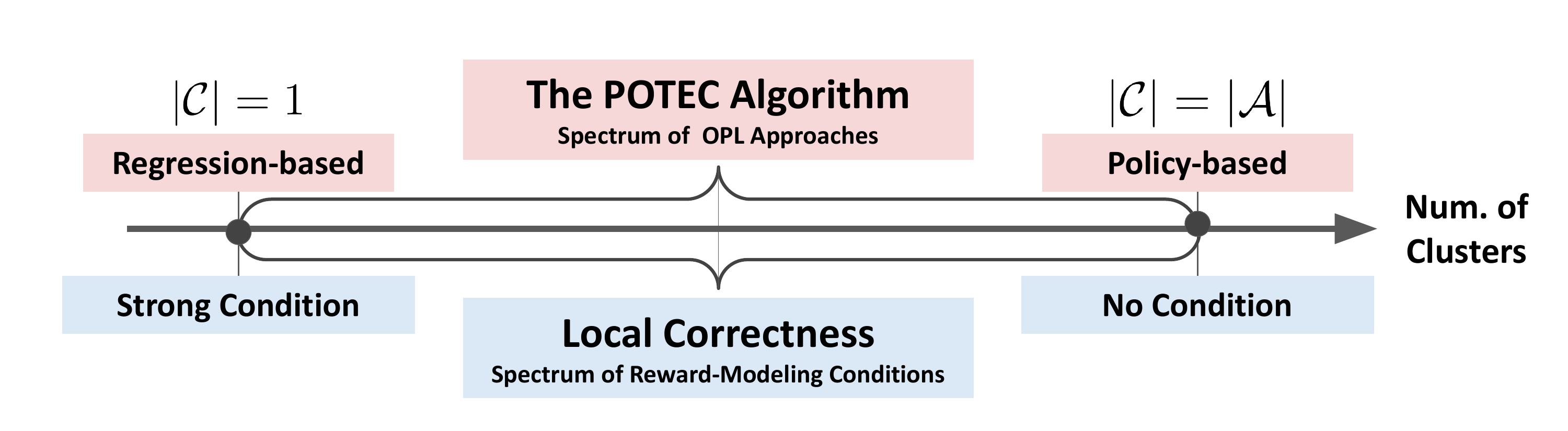}
\vspace*{-3mm}
\caption{The POTEC algorithm and local correctness condition generalize policy- and regression-based approaches and their respective conditions about the reward function ($q(x,a)$) estimation.} \label{fig:generalization}
\raggedright
\end{figure*}

\begin{algorithm*}[t]
\caption{The POTEC Algorithm} \label{algo:potec}
\begin{algorithmic}[1]
\REQUIRE logged bandit data $\calD$, logging policy $\pi_0$, action clustering function $c_{a}$.
\ENSURE 1st-stage (policy-based) policy $\firstpolicy$ and 2nd-stage (regression-based) policy $\secondpolicy$
\STATE Perform pairwise regression and obtain $\hat{h}_{\psi}(x,a)$ as in Eq.~\eqref{eq:bias-minimization}, which works as the 2nd-stage policy as in Eq.~\eqref{eq:2nd-stage-policy} and also as a part of the regression model to help train the 1st-stage policy via the POTEC gradient estimator
\STATE Regress the reward residual from pairwise regression and obtain $\hat{g}_{\omega}(x,c)$ as in Eq.~\eqref{eq:variance-minimization}
\STATE Perform policy-based learning of the 1st-stage policy based on the POTEC gradient estimator in Eq.~\eqref{eq:potec-pg}
\end{algorithmic}
\end{algorithm*}
\subsection{Training the 2nd-Stage Policy $\secondpolicy$} \label{sec:2nd-stage-policy-training}
We have thus far developed a policy-based approach for learning an effective cluster selection (1st-stage) policy via the POTEC gradient estimator. The remaining objective is to identify the optimal actions, given a cluster selected by the 1st-stage policy. In essence, we should be able to simply use the pairwise regression model $\hat{h}_{\psi}$ from the previous section to establish the 2nd-stage policy $\secondpolicy$, because $\hat{h}_{\psi}$ is already optimized towards estimating the relative value differences of actions within each action cluster (i.e., local correctness). Specifically, we suggest constructing a conditional action selection (2nd-stage) policy based on $\hat{h}_{\psi}$ as
\begin{align}
    \secondpolicy (a \,|\, x,c) :=  \left\{
    \begin{array}{ll}
        1 & (a = \argmax_{a':c_{a'}=c} \, \hat{h}_{\psi}(x,a')) \\
        0 & (\text{otherwise})
    \end{array}
    \right. \label{eq:2nd-stage-policy}
\end{align}
which implies that the 2nd-stage policy selects the action with the highest value of the pairwise regression function $\hat{h}_{\psi}$ within the already sampled cluster $c$. This action selection procedure is justified since we have learned the function $\hat{h}_{\psi}$ so that it can estimate the relative value difference of the actions given a cluster in the bias minimization step (Eq.~\eqref{eq:bias-minimization}). In an ideal scenario where Condition~\ref{assumption:local_correctness} holds true, our 2nd-stage policy achieves optimal action selection. In our experiments, we will demonstrate that our overall policy $\overallpolicy$ outperforms existing approaches by a considerable margin even with a learned 2nd-stage policy that may not perfectly satisfy local correctness.

\subsection{The Overall POTEC Algorithm}  \label{sec:algorithm}
Algorithm~\ref{algo:potec} describes the overall procedure of our POTEC algorithm. It first performs the bias and variance minimization steps to obtain $\hat{h}_{\psi}$ and $\hat{g}_{\omega}$ where $\hat{h}_{\psi}$ forms the 2nd-stage policy (as in Eq.~\eqref{eq:2nd-stage-policy}). Then, we train the 1st-stage policy $\firstpolicy$ based on the POTEC gradient estimator, which is based on cluster importance weighting and a learned regression model $\hat{f}_{\psi,\omega}(x,a) = \hat{g}_{\omega}(x,c_a) + \hat{h}_{\psi}(x,a)$.

It is worth mentioning that POTEC and its associated local correctness condition generalize typical OPL approaches, i.e., policy-based and regression-based, as depicted in Figure~\ref{fig:generalization}. That is, when there is only one action cluster ($|\calC|=1$), the 2nd-stage policy of POTEC needs to choose the best action in the entire action space, which can be seen as a reduction to the regression-based approach. Moreover, in this case, the local correctness condition becomes relatively stringent (since all actions are grouped into the same cluster), which is also akin to the typical condition of the regression-based approach, i.e., globally accurate estimation of the reward function. On the other hand, when the cluster space is equivalent to the original action space ($\calC=\calA$), the 1st-stage policy selects an action from the original action space, akin to the policy-based approach. In this scenario, the local correctness condition imposes no specific requirements, as each action cluster contains only one unique action. This absence of requirements aligns with the policy-based approach, which does not necessitate specific conditions for reward function estimation to produce an unbiased gradient. Thus, POTEC and local correctness encompass the full spectrum of existing OPL approaches and respective reward-modeling conditions (Figure~\ref{fig:generalization}). As a strict generalization, POTEC offers the potential to enhance both approaches with a good selection of the number of clusters, as the following section empirically demonstrates.

\section{Empirical Evaluation} \label{sec:empirical}
We first evaluate POTEC on synthetic data with the ground-truth cluster information to identify the situations where it enables more effective OPL. We then assess the real-world applicability of POTEC with learned clusters on two extreme classification datasets using the standard supervised-to-bandit methodology \cite{dudik2011doubly,su2019cab}. Our experiments are conducted using the \textit{OpenBanditPipeline} (OBP)\footnote{\href{https://github.com/st-tech/zr-obp}{\blue{https://github.com/st-tech/zr-obp}}}, an open-source software for OPE provided by~\citep{saito2021open}. 

\begin{figure*}[t]
\centering
\begin{minipage}{\hsize}
    \begin{center}
        \includegraphics[clip, width=15cm]{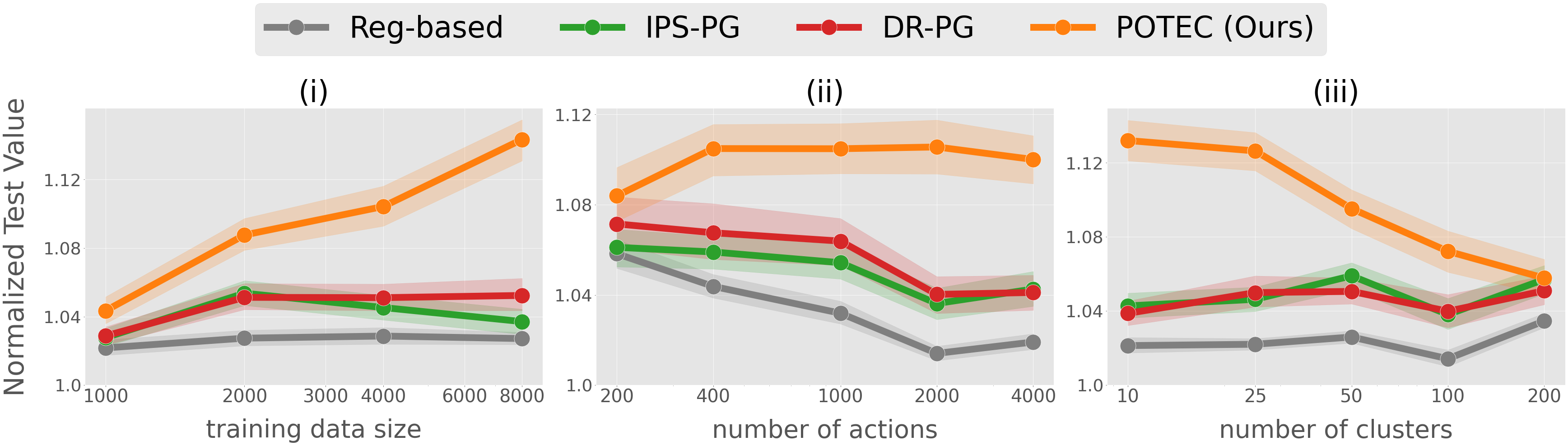}
    \end{center}
    \vspace{-2mm}
    \caption{Comparing the test policy value (normalized by $V(\pi_0)$) of the OPL methods, with varying \textbf{(i)} training data sizes, \textbf{(ii)} numbers of actions, and \textbf{(iii)} numbers of (true) clusters, in the synthetic experiment.}
    \label{fig:synthetic}
\end{minipage}
\end{figure*}
\subsection{Synthetic Data} \label{sec:synthetic}
We create synthetic datasets to be able to compare the policy learning algorithms based on their ground-truth value. Specifically, we first sample 10-dimensional context vectors $x$ and features of the actions from the standard normal distribution. We then form (true) action clusters based on the action features. We synthesize the expected reward function as $q(x,a) = g(x,c_a) + h_{c_a}(x,a)$ where $g(\cdot,\cdot)$ and $h_{\cdot}(\cdot,\cdot)$ define the values of cluster and individual action respectively, as detailed in Appendix~\ref{app:experiment}. Finally, we sample binary reward $r$ from a Bernoulli distribution with mean $q(x,a)$.

\begin{figure}[t]
\centering
\includegraphics[width=0.8\linewidth]{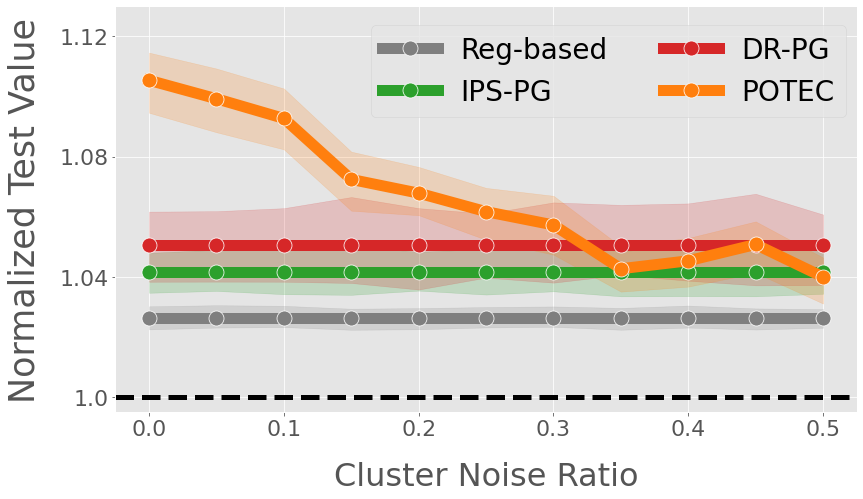}  \vspace{-2mm}
\caption{Comparing the test policy value (normalized by $V(\pi_0)$) of the OPL methods under varying cluster noise ratios.} \label{fig:ablation}
\end{figure}
\paragraph{Baselines:}
We compare POTEC with the regression-based method (Reg-based), IPS-PG (Eq.~\eqref{eq:is-pg}), and DR-PG (Eq.~\eqref{eq:dr-pg}). We use a neural network with 3 hidden layers to parameterize the policy $\pi_{\theta}$, $\hat{q}(x,a)$ for DR-PG and Reg-based, and $(\hat{h}_{\psi}, \hat{g}_{\omega})$ for POTEC. We also apply the variance reduction technique proposed by~\citet{lopez2021learning} to IPS-PG and DR-PG. Note that we use a fixed set of hyper-parameters for POTEC while we tune the hyper-parameters of only the baselines based on the true policy value in the test set, which gives the baselines an unfair advantage.

\begin{figure}[t]
\centering
\includegraphics[width=1\linewidth]{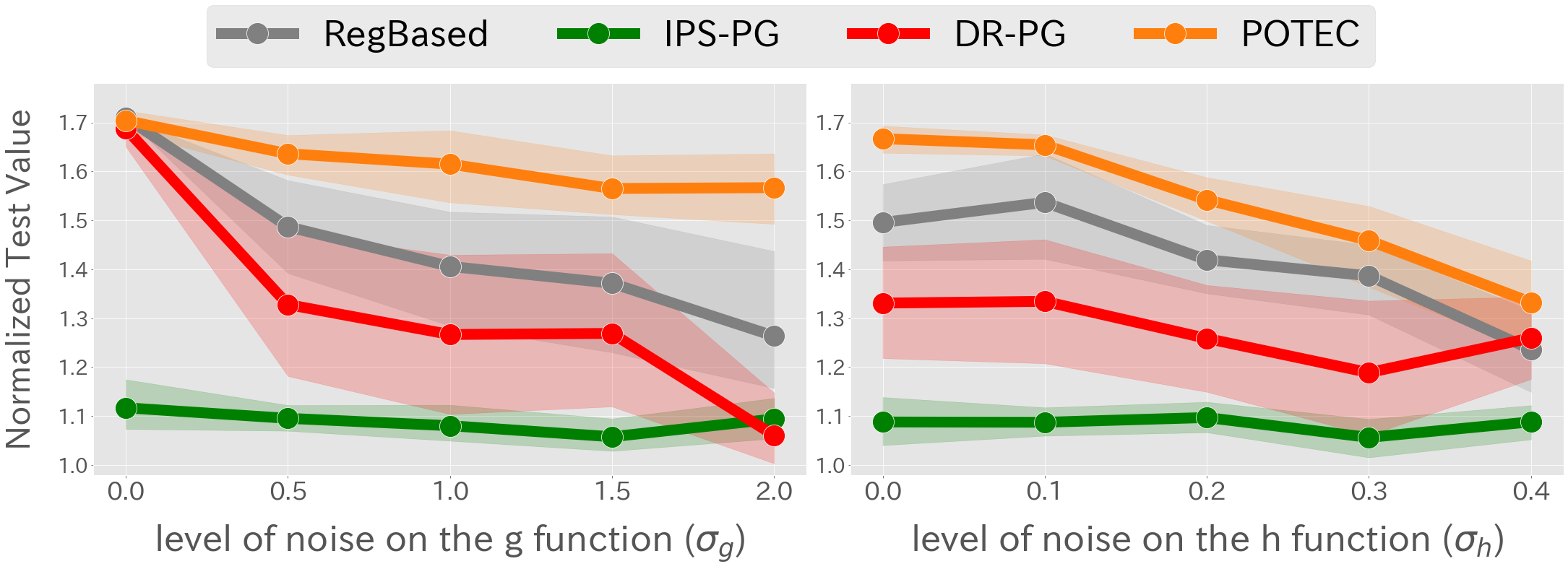}  \vspace{-5mm}
\caption{Comparing the test policy value (normalized by $V(\pi_0)$) of the OPL methods under varying accuracies of $\hat{q}$ and $\hat{f}$.} \label{fig:reward-model-noise}
\end{figure}
\begin{figure*}[t]
\centering
\begin{minipage}{\hsize}
    \begin{center}
        \includegraphics[clip, width=14cm]{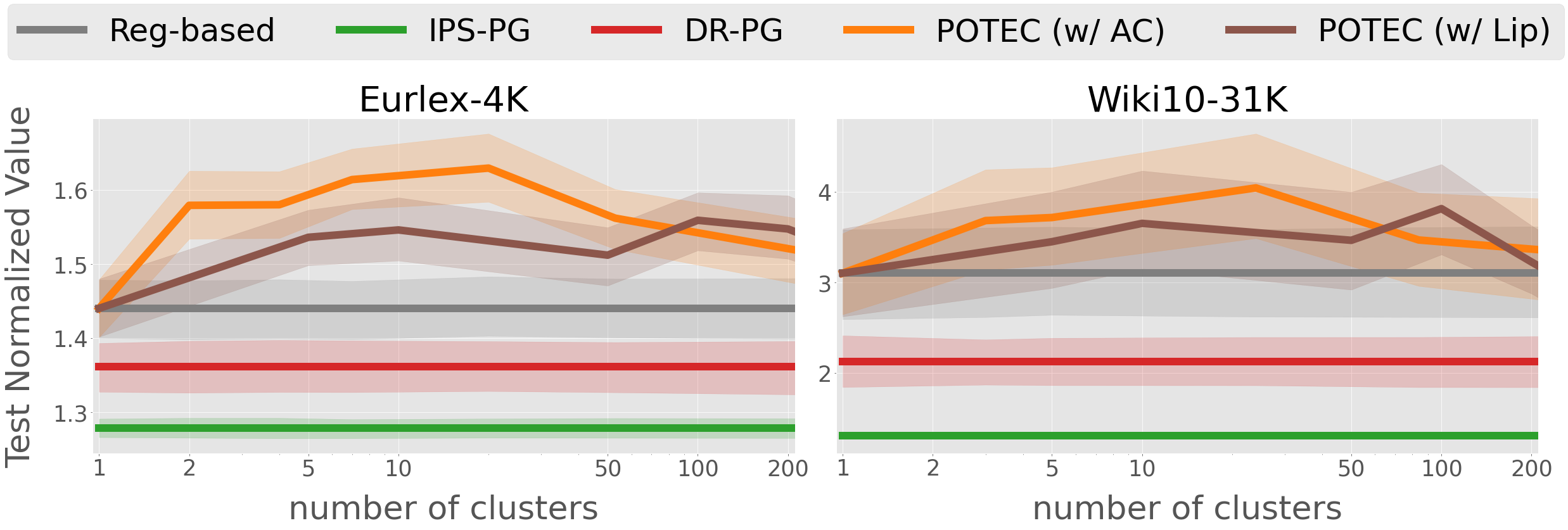}
    \end{center}
    \vspace{-2mm}
    \caption{Comparing the test policy value (normalized by $V(\pi_0)$) of the OPL methods, with varying numbers of clusters (hyper-parameter of POTEC) on the EUR-Lex 4K and Wiki10-31K datasets.}
    \label{fig:real}
\end{minipage}
\end{figure*}
\paragraph{Results:} Figure~\ref{fig:synthetic} shows the policy values of the OPL methods on test data obtained from 100 simulations with varying random seeds. Note that we employ default experiment parameters of $n=4,000$, $|\calA|=2,000$, and $|\calC|=30$.

First, in all situations, POTEC provides significant improvements in policy value over the baseline methods, even though they possess an unfair advantage in terms of hyperparameter tuning. Specifically, in Figure~\ref{fig:synthetic} (i), we can see that POTEC performs increasingly better with increasing sample sizes while the baseline methods do not improve. This suggests that POTEC is more sample-efficient in large action spaces, while the baseline methods need even larger datasets to be effective. Next, in Figure~\ref{fig:synthetic} (ii), we vary the number of actions ($|\calA|$) to investigate the robustness to growing action spaces. We can see that POTEC performs consistently even with growing action spaces as long as the cluster space does not grow, while the performance of the baseline methods worsens clearly for larger numbers of actions. Finally, Figure~\ref{fig:synthetic} (iii) evaluates POTEC as we increase the number of (true) clusters while keeping the number of actions fixed. The figure shows that the advantage of POTEC becomes largest when the cluster effect can be captured by a small number of underlying clusters; however, even for synthetic data with $|\calC|=200$ clusters, POTEC remains highly competitive with the baselines (note that the baselines have an unfair advantage in hyperparameter tuning). Appendix~\ref{app:experiment} reports more experiment results showing that POTEC performs consistently better under varying logging policies and the violation of full support.

In Figure~\ref{fig:ablation}, we report the result of an ablation study under the default setup ($n=4,000$, $|\calA|=2,000$, and $|\calC|=30$) where we add some noise to the clusters by flipping the true cluster membership of actions with some given probability (cluster noise ratio). It shows that POTEC is particularly powerful with accurate cluster information, but it remains superior to the baselines even when 30\% of the cluster information is perturbed. We can also see that POTEC performs similarly to the policy-based baselines even when about half of the cluster information is not accurate.

In Figure~\ref{fig:reward-model-noise}, we compare varying accuracies of the regression model ($\hat{q}$ for DR-PG and Reg-based, and $\hat{f}$ for POTEC). For this study, we define the (synthetic) regression model as $\hat{q}(x,a)=\hat{f}(x,a) = (g(x,c_a) + \epsilon_{c_a})+(h_{c_a}(x,a) + \epsilon_a)$ where $\epsilon_{c},\epsilon_a$ are Gaussian noises with different standard deviations $\sigma_c$ and $\sigma_a$. In Figure~\ref{fig:reward-model-noise} (left), we vary $\sigma_c$ with $\sigma_a$ being fixed at $0.0$, while in Figure~\ref{fig:reward-model-noise} (right), we vary $\sigma_a$  with $\sigma_c$ being fixed at $0.3$. This approach allows us to investigate the impact of errors in estimating cluster and action effects (the $g$ and $h$ functions) on different methods. First, Figure~\ref{fig:reward-model-noise} (left) shows that POTEC is not significantly affected by noise in the cluster value ($g$) and remains effective throughout, whereas the Reg-based method deteriorates substantially. This is attributed to the fact that the POTEC gradient estimator for the first-stage remains unbiased, and the effectiveness of the second-stage policy is maintained irrespective of the noise in the cluster value. Secondly, Figure~\ref{fig:reward-model-noise} (right) reveals that noise in the action effect ($h$) impacts both POTEC and Reg-based methods. However, POTEC exhibits greater robustness, as it does not solely rely on the regression model to learn the overall policy. These results demonstrate POTEC's robustness against reward estimation errors. The results also highlight the benefits of employing pairwise regression to directly minimize errors against the action effect that has larger adverse effects on the effectiveness of POTEC.

\subsection{Real-World Data} \label{sec:real}
To assess the real-world applicability of POTEC, we now evaluate it on the EUR-Lex 4K and Wiki10-31K datasets, extreme classification data with several thousands of labels (actions) provided in the Extreme Classification Repository~\citep{bhatia2016extreme}.

To perform an OPL experiment, we convert the extreme classification datasets with $L$ labels into contextual bandit datasets with the same number of actions. Table~\ref{tab:data_stats} in Appendix~\ref{app:experiment} shows the statistics of the real-world datasets such as the number of datapoints and actions. We consider stochastic rewards with the expected reward function of the form: $q(x, a)= (1-\eta_a) \mathbb{I}\{\text {if $a$ has a positive label}\} + \eta_a \mathbb{I}\{\text {if $a$ has a negative label}\}$ where $\mathbb{I}\{\cdot\}$ is the indicator function and $\eta_a$ is a noise parameter sampled separately for each action $a$ from a uniform distribution with range $[0,0.1]$. We then sample the reward from a normal distribution with mean $q(x,a)$ and standard deviation $\sigma = 0.05$.

We define the logging policy $\pi_0$ by applying the softmax function to an estimated reward function $\hat{q}(x,a)$, which is obtained by a matrix factorization model and is different from the estimated reward function used in POTEC, DR-PG, and the Reg-based method. More details of the real-world experiment setup can be found in Appendix~\ref{app:experiment}.

\paragraph{Results:}
We evaluate POTEC against IPS-PG, DR-PG, and Reg-based under varying numbers of clusters to evaluate POTEC's robustness to the choice of this key hyper-parameter. We optimize the hyperparameters of POTEC and the baselines based on the ground-truth policy value in the validation set, and the effectiveness of the OPL methods is evaluated on the test set. For POTEC, we evaluate it with two types of clustering methods to investigate its robustness to the ways the clustering is performed. The first method is through learning an action embedding via Lipschitz regularization (Lip) recently proposed for improving OPE in large action spaces~\cite{peng2023offline}. The second method is to apply Agglomerative clustering (AC) implemented in scikit-learn~\citep{pedregosa2011scikit} to the full-information labels, which provides an even more accurate clustering by leveraging the true reward correlation. Note that we perform a conventional reward regression rather than the two-step regression for POTEC due to insufficient pairwise data in these specific datasets.

Figure~\ref{fig:real} presents the test policy value (normalized by $V(\pi_0)$) of the OPL methods with varying numbers of clusters (the hyper-parameter of POTEC) on Eurlex-4K (left) and Wiki10-31K (right). Note that the baseline methods do not depend on action clusters, leading to flat lines. The figure reveals that POTEC with both clustering methods outperforms all baseline methods on both datasets given a moderate number of clusters (2 to 100) indicating its potential for real-world applications even with action clustering learned only from observable logged data (i.e., POTEC w/ Lip). We can also see that POTEC with a more accurate clustering (i.e., POTEC w/ AC) slightly outperforms POTEC w/ Lip, implying an even better potential of POTEC with a more refined clustering procedure.

\section{Conclusion and Future Work} \label{sec:conclusion}
This work introduces a novel two-stage OPL procedure called POTEC, which is particularly advantageous in large action spaces. POTEC learns the first-stage cluster-selection policy via a new policy gradient estimator, which is unbiased under local correctness and has substantially lower variance. The second-stage action-selection policy is learned through pairwise reward regression, offering greater robustness to bias compared to traditional regression-based approaches. We also provide an intriguing interpretation of POTEC and local correctness as a full spectrum of existing approaches in OPL and respective reward-modeling conditions.

Our findings give rise to valuable directions for future studies. For example, even though we have empirically demonstrated that POTEC outperforms existing OPL methods with some heuristic action clustering on real-world data, it would be valuable to consider a more refined clustering method such as an iterative procedure to optimize the clustering and the regression model simultaneously to satisfy local correctness better. Extension of POTEC to offline reinforcement learning and large language models beyond generic contextual bandits is also an interesting future direction.

\section*{Acknowledgements}
This research was supported in part by NSF Awards IIS-1901168 and IIS-2008139. Yuta Saito was supported by Funai Overseas Scholarship. All content represents the opinion of the authors, which is not necessarily shared or endorsed by their respective employers and/or sponsors.

\bibliography{main.bbl}

\begin{thebibliography}{51}
\providecommand{\natexlab}[1]{#1}
\providecommand{\url}[1]{\texttt{#1}}
\expandafter\ifx\csname urlstyle\endcsname\relax
  \providecommand{\doi}[1]{doi: #1}\else
  \providecommand{\doi}{doi: \begingroup \urlstyle{rm}\Url}\fi

\bibitem[Agrawal \& Goyal(2013)Agrawal and Goyal]{agrawal2013thompson}
Agrawal, S. and Goyal, N.
\newblock Thompson sampling for contextual bandits with linear payoffs.
\newblock In \emph{International Conference on Machine Learning}, pp.\  127--135. PMLR, 2013.

\bibitem[Athey et~al.(2019)Athey, Chetty, Imbens, and Kang]{athey2019surrogate}
Athey, S., Chetty, R., Imbens, G.~W., and Kang, H.
\newblock The surrogate index: Combining short-term proxies to estimate long-term treatment effects more rapidly and precisely.
\newblock Technical report, National Bureau of Economic Research, 2019.

\bibitem[Athey et~al.(2020)Athey, Chetty, and Imbens]{athey2020combining}
Athey, S., Chetty, R., and Imbens, G.
\newblock Combining experimental and observational data to estimate treatment effects on long term outcomes.
\newblock \emph{arXiv preprint arXiv:2006.09676}, 2020.

\bibitem[Bhatia et~al.(2016)Bhatia, Dahiya, Jain, Kar, Mittal, Prabhu, and Varma]{bhatia2016extreme}
Bhatia, K., Dahiya, K., Jain, H., Kar, P., Mittal, A., Prabhu, Y., and Varma, M.
\newblock The extreme classification repository: Multi-label datasets and code, 2016.
\newblock URL \url{http://manikvarma.org/downloads/XC/XMLRepository.html}.

\bibitem[Chandak et~al.(2019)Chandak, Theocharous, Kostas, Jordan, and Thomas]{chandak2019learning}
Chandak, Y., Theocharous, G., Kostas, J., Jordan, S., and Thomas, P.
\newblock Learning action representations for reinforcement learning.
\newblock In \emph{International Conference on Machine Learning}, pp.\  941--950. PMLR, 2019.

\bibitem[Chen \& Ritzwoller(2021)Chen and Ritzwoller]{chen2021semiparametric}
Chen, J. and Ritzwoller, D.~M.
\newblock Semiparametric estimation of long-term treatment effects.
\newblock \emph{arXiv preprint arXiv:2107.14405}, 2021.

\bibitem[Dud{\'\i}k et~al.(2011)Dud{\'\i}k, Langford, and Li]{dudik2011doubly}
Dud{\'\i}k, M., Langford, J., and Li, L.
\newblock Doubly robust policy evaluation and learning.
\newblock In \emph{Proceedings of the 28th International Conference on International Conference on Machine Learning}, pp.\  1097--1104, 2011.

\bibitem[Dud{\'\i}k et~al.(2014)Dud{\'\i}k, Erhan, Langford, and Li]{dudik2014doubly}
Dud{\'\i}k, M., Erhan, D., Langford, J., and Li, L.
\newblock Doubly robust policy evaluation and optimization.
\newblock \emph{Statistical Science}, 29\penalty0 (4):\penalty0 485--511, 2014.

\bibitem[Farajtabar et~al.(2018)Farajtabar, Chow, and Ghavamzadeh]{farajtabar2018more}
Farajtabar, M., Chow, Y., and Ghavamzadeh, M.
\newblock More robust doubly robust off-policy evaluation.
\newblock In \emph{Proceedings of the 35th International Conference on Machine Learning}, volume~80, pp.\  1447--1456. PMLR, 2018.

\bibitem[Felicioni et~al.(2022)Felicioni, Ferrari~Dacrema, Restelli, and Cremonesi]{felicioni2022off}
Felicioni, N., Ferrari~Dacrema, M., Restelli, M., and Cremonesi, P.
\newblock Off-policy evaluation with deficient support using side information.
\newblock \emph{Advances in Neural Information Processing Systems}, 35, 2022.

\bibitem[Gu et~al.(2022)Gu, Zhao, Chen, Li, Hao, and An]{gu2022learning}
Gu, P., Zhao, M., Chen, C., Li, D., Hao, J., and An, B.
\newblock Learning pseudometric-based action representations for offline reinforcement learning.
\newblock In \emph{International Conference on Machine Learning}, pp.\  7902--7918. PMLR, 2022.

\bibitem[Jeunen \& Goethals(2021)Jeunen and Goethals]{jeunen2021pessimistic}
Jeunen, O. and Goethals, B.
\newblock Pessimistic reward models for off-policy learning in recommendation.
\newblock In \emph{Proceedings of the 15th ACM Conference on Recommender Systems}, pp.\  63--74, 2021.

\bibitem[Jiang \& Li(2016)Jiang and Li]{jiang2016doubly}
Jiang, N. and Li, L.
\newblock Doubly robust off-policy value evaluation for reinforcement learning.
\newblock In \emph{Proceedings of the 33rd International Conference on Machine Learning}, volume~48, pp.\  652--661. PMLR, 2016.

\bibitem[Joachims et~al.(2018)Joachims, Swaminathan, and de~Rijke]{joachims2018deep}
Joachims, T., Swaminathan, A., and de~Rijke, M.
\newblock Deep learning with logged bandit feedback.
\newblock In \emph{International Conference on Learning Representations}, 2018.

\bibitem[Kallus \& Uehara(2020)Kallus and Uehara]{kallus2020double}
Kallus, N. and Uehara, M.
\newblock Double reinforcement learning for efficient off-policy evaluation in markov decision processes.
\newblock \emph{J. Mach. Learn. Res.}, 21:\penalty0 167--1, 2020.

\bibitem[Kallus et~al.(2021)Kallus, Saito, and Uehara]{kallus2020optimal}
Kallus, N., Saito, Y., and Uehara, M.
\newblock Optimal off-policy evaluation from multiple logging policies.
\newblock In \emph{Proceedings of the 38th International Conference on Machine Learning}, volume 139, pp.\  5247--5256. PMLR, 2021.

\bibitem[Kingma \& Ba(2014)Kingma and Ba]{kingma2014adam}
Kingma, D.~P. and Ba, J.
\newblock Adam: A method for stochastic optimization.
\newblock \emph{arXiv preprint arXiv:1412.6980}, 2014.

\bibitem[Kiyohara et~al.(2022)Kiyohara, Saito, Matsuhiro, Narita, Shimizu, and Yamamoto]{kiyohara2022doubly}
Kiyohara, H., Saito, Y., Matsuhiro, T., Narita, Y., Shimizu, N., and Yamamoto, Y.
\newblock Doubly robust off-policy evaluation for ranking policies under the cascade behavior model.
\newblock In \emph{Proceedings of the 15th International Conference on Web Search and Data Mining}, 2022.

\bibitem[Kiyohara et~al.(2023)Kiyohara, Uehara, Narita, Shimizu, Yamamoto, and Saito]{kiyohara2023off}
Kiyohara, H., Uehara, M., Narita, Y., Shimizu, N., Yamamoto, Y., and Saito, Y.
\newblock Off-policy evaluation of ranking policies under diverse user behavior.
\newblock In \emph{Proceedings of the 29th ACM SIGKDD Conference on Knowledge Discovery and Data Mining}, pp.\  1154--1163, 2023.

\bibitem[Kiyohara et~al.(2024{\natexlab{a}})Kiyohara, Kishimoto, Kawakami, Kobayashi, Nakata, and Saito]{kiyohara2024towards}
Kiyohara, H., Kishimoto, R., Kawakami, K., Kobayashi, K., Nakata, K., and Saito, Y.
\newblock Towards assessing and benchmarking risk-return tradeoff of off-policy evaluation.
\newblock In \emph{International Conference on Learning Representations}, 2024{\natexlab{a}}.

\bibitem[Kiyohara et~al.(2024{\natexlab{b}})Kiyohara, Masahiro, and Saito]{kiyohara2024off}
Kiyohara, H., Masahiro, N., and Saito, Y.
\newblock Off-policy evaluation of slate bandit policies via optimizing abstraction.
\newblock In \emph{Proceedings of the ACM Web Conference 2024}, 2024{\natexlab{b}}.

\bibitem[Lattimore \& Szepesv{\'a}ri(2020)Lattimore and Szepesv{\'a}ri]{lattimore2020bandit}
Lattimore, T. and Szepesv{\'a}ri, C.
\newblock \emph{Bandit algorithms}.
\newblock Cambridge University Press, 2020.

\bibitem[Lee et~al.(2022)Lee, Arbour, and Theocharous]{lee2022off}
Lee, J.~J., Arbour, D., and Theocharous, G.
\newblock Off-policy evaluation in embedded spaces.
\newblock \emph{arXiv preprint arXiv:2203.02807}, 2022.

\bibitem[Li et~al.(2010)Li, Chu, Langford, and Schapire]{li2010contextual}
Li, L., Chu, W., Langford, J., and Schapire, R.~E.
\newblock A contextual-bandit approach to personalized news article recommendation.
\newblock In \emph{Proceedings of the 19th international conference on World wide web}, pp.\  661--670, 2010.

\bibitem[Liang \& Vlassis(2022)Liang and Vlassis]{liang2022local}
Liang, D. and Vlassis, N.
\newblock Local policy improvement for recommender systems.
\newblock \emph{arXiv preprint arXiv:2212.11431}, 2022.

\bibitem[Liu et~al.(2018)Liu, Li, Tang, and Zhou]{liu2018breaking}
Liu, Q., Li, L., Tang, Z., and Zhou, D.
\newblock Breaking the curse of horizon: infinite-horizon off-policy estimation.
\newblock In \emph{Proceedings of the 32nd International Conference on Neural Information Processing Systems}, pp.\  5361--5371, 2018.

\bibitem[Liu et~al.(2020)Liu, Bacon, and Brunskill]{liu2020understanding}
Liu, Y., Bacon, P.-L., and Brunskill, E.
\newblock Understanding the curse of horizon in off-policy evaluation via conditional importance sampling.
\newblock In \emph{International Conference on Machine Learning}, pp.\  6184--6193. PMLR, 2020.

\bibitem[London \& Sandler(2019)London and Sandler]{london2019bayesian}
London, B. and Sandler, T.
\newblock Bayesian counterfactual risk minimization.
\newblock In \emph{International Conference on Machine Learning}, pp.\  4125--4133. PMLR, 2019.

\bibitem[Lopez et~al.(2021)Lopez, Dhillon, and Jordan]{lopez2021learning}
Lopez, R., Dhillon, I.~S., and Jordan, M.~I.
\newblock Learning from extreme bandit feedback.
\newblock \emph{Proc. Association for the Advancement of Artificial Intelligence}, 2021.

\bibitem[Ma et~al.(2019)Ma, Wang, and Narayanaswamy]{ma2019imitation}
Ma, Y., Wang, Y.-X., and Narayanaswamy, B.
\newblock Imitation-regularized offline learning.
\newblock In \emph{The 22nd International Conference on Artificial Intelligence and Statistics}, pp.\  2956--2965. PMLR, 2019.

\bibitem[Metelli et~al.(2021)Metelli, Russo, and Restelli]{metelli2021subgaussian}
Metelli, A.~M., Russo, A., and Restelli, M.
\newblock Subgaussian and differentiable importance sampling for off-policy evaluation and learning.
\newblock \emph{Advances in Neural Information Processing Systems}, 34, 2021.

\bibitem[Pedregosa et~al.(2011)Pedregosa, Varoquaux, Gramfort, Michel, Thirion, Grisel, Blondel, Prettenhofer, Weiss, Dubourg, Vanderplas, Passos, Cournapeau, Brucher, Perrot, and {{\'E}}douard Duchesnay]{pedregosa2011scikit}
Pedregosa, F., Varoquaux, G., Gramfort, A., Michel, V., Thirion, B., Grisel, O., Blondel, M., Prettenhofer, P., Weiss, R., Dubourg, V., Vanderplas, J., Passos, A., Cournapeau, D., Brucher, M., Perrot, M., and {{\'E}}douard Duchesnay.
\newblock Scikit-learn: Machine learning in python.
\newblock \emph{Journal of Machine Learning Research}, 12:\penalty0 2825--2830, 2011.

\bibitem[Peng et~al.(2023)Peng, Zou, Liu, Li, Jiang, Pei, and Cui]{peng2023offline}
Peng, J., Zou, H., Liu, J., Li, S., Jiang, Y., Pei, J., and Cui, P.
\newblock Offline policy evaluation in large action spaces via outcome-oriented action grouping.
\newblock In \emph{Proceedings of the ACM Web Conference 2023}, pp.\  1220--1230, 2023.

\bibitem[Sachdeva et~al.(2020)Sachdeva, Su, and Joachims]{sachdeva2020off}
Sachdeva, N., Su, Y., and Joachims, T.
\newblock Off-policy bandits with deficient support.
\newblock In \emph{Proceedings of the 26th ACM SIGKDD International Conference on Knowledge Discovery and Data Mining}, pp.\  965--975, 2020.

\bibitem[Sachdeva et~al.(2023)Sachdeva, Wang, Liang, Kallus, and McAuley]{sachdeva2023off}
Sachdeva, N., Wang, L., Liang, D., Kallus, N., and McAuley, J.
\newblock Off-policy evaluation for large action spaces via policy convolution.
\newblock \emph{arXiv preprint arXiv:2310.15433}, 2023.

\bibitem[Saito \& Joachims(2021)Saito and Joachims]{saito2021counterfactual}
Saito, Y. and Joachims, T.
\newblock Counterfactual learning and evaluation for recommender systems: Foundations, implementations, and recent advances.
\newblock In \emph{Proceedings of the 15th ACM Conference on Recommender Systems}, pp.\  828–830, 2021.

\bibitem[Saito \& Joachims(2022)Saito and Joachims]{saito2022off}
Saito, Y. and Joachims, T.
\newblock Off-policy evaluation for large action spaces via embeddings.
\newblock In \emph{International Conference on Machine Learning}, pp.\  19089--19122. PMLR, 2022.

\bibitem[Saito et~al.(2021{\natexlab{a}})Saito, Aihara, Matsutani, and Narita]{saito2021open}
Saito, Y., Aihara, S., Matsutani, M., and Narita, Y.
\newblock Open bandit dataset and pipeline: Towards realistic and reproducible off-policy evaluation.
\newblock In \emph{Thirty-fifth Conference on Neural Information Processing Systems Datasets and Benchmarks Track}, 2021{\natexlab{a}}.

\bibitem[Saito et~al.(2021{\natexlab{b}})Saito, Udagawa, Kiyohara, Mogi, Narita, and Tateno]{saito2021evaluating}
Saito, Y., Udagawa, T., Kiyohara, H., Mogi, K., Narita, Y., and Tateno, K.
\newblock Evaluating the robustness of off-policy evaluation.
\newblock In \emph{Proceedings of the 15th ACM Conference on Recommender Systems}, pp.\  114–123, 2021{\natexlab{b}}.

\bibitem[Saito et~al.(2023)Saito, Qingyang, and Joachims]{saito2023off}
Saito, Y., Qingyang, R., and Joachims, T.
\newblock Off-policy evaluation for large action spaces via conjunct effect modeling.
\newblock In \emph{International Conference on Machine Learning}, pp.\  29734--29759. PMLR, 2023.

\bibitem[Su et~al.(2019)Su, Wang, Santacatterina, and Joachims]{su2019cab}
Su, Y., Wang, L., Santacatterina, M., and Joachims, T.
\newblock Cab: Continuous adaptive blending for policy evaluation and learning.
\newblock In \emph{International Conference on Machine Learning}, volume~84, pp.\  6005--6014, 2019.

\bibitem[Su et~al.(2020{\natexlab{a}})Su, Dimakopoulou, Krishnamurthy, and Dud{\'\i}k]{su2020doubly}
Su, Y., Dimakopoulou, M., Krishnamurthy, A., and Dud{\'\i}k, M.
\newblock Doubly robust off-policy evaluation with shrinkage.
\newblock In \emph{Proceedings of the 37th International Conference on Machine Learning}, volume 119, pp.\  9167--9176. PMLR, 2020{\natexlab{a}}.

\bibitem[Su et~al.(2020{\natexlab{b}})Su, Srinath, and Krishnamurthy]{su2020adaptive}
Su, Y., Srinath, P., and Krishnamurthy, A.
\newblock Adaptive estimator selection for off-policy evaluation.
\newblock In \emph{International Conference on Machine Learning}, pp.\  9196--9205. PMLR, 2020{\natexlab{b}}.

\bibitem[Swaminathan \& Joachims(2015{\natexlab{a}})Swaminathan and Joachims]{swaminathan2015batch}
Swaminathan, A. and Joachims, T.
\newblock Batch learning from logged bandit feedback through counterfactual risk minimization.
\newblock \emph{The Journal of Machine Learning Research}, 16\penalty0 (1):\penalty0 1731--1755, 2015{\natexlab{a}}.

\bibitem[Swaminathan \& Joachims(2015{\natexlab{b}})Swaminathan and Joachims]{swaminathan2015counterfactual}
Swaminathan, A. and Joachims, T.
\newblock Counterfactual risk minimization: Learning from logged bandit feedback.
\newblock In \emph{International Conference on Machine Learning}, pp.\  814--823. PMLR, 2015{\natexlab{b}}.

\bibitem[Swaminathan \& Joachims(2015{\natexlab{c}})Swaminathan and Joachims]{swaminathan2015self}
Swaminathan, A. and Joachims, T.
\newblock The self-normalized estimator for counterfactual learning.
\newblock \emph{Advances in Neural Information Processing Systems}, 28, 2015{\natexlab{c}}.

\bibitem[Thomas \& Brunskill(2016)Thomas and Brunskill]{thomas2016data}
Thomas, P. and Brunskill, E.
\newblock Data-efficient off-policy policy evaluation for reinforcement learning.
\newblock In \emph{Proceedings of the 33rd International Conference on Machine Learning}, volume~48, pp.\  2139--2148. PMLR, 2016.

\bibitem[Udagawa et~al.(2023)Udagawa, Kiyohara, Narita, Saito, and Tateno]{udagawa2023policy}
Udagawa, T., Kiyohara, H., Narita, Y., Saito, Y., and Tateno, K.
\newblock Policy-adaptive estimator selection for off-policy evaluation.
\newblock In \emph{Proceedings of the AAAI Conference on Artificial Intelligence}, volume~36, 2023.

\bibitem[Voloshin et~al.(2019)Voloshin, Le, Jiang, and Yue]{voloshin2019empirical}
Voloshin, C., Le, H.~M., Jiang, N., and Yue, Y.
\newblock Empirical study of off-policy policy evaluation for reinforcement learning.
\newblock \emph{arXiv preprint arXiv:1911.06854}, 2019.

\bibitem[Wang et~al.(2017)Wang, Agarwal, and Dud{\i}k]{wang2017optimal}
Wang, Y.-X., Agarwal, A., and Dud{\i}k, M.
\newblock Optimal and adaptive off-policy evaluation in contextual bandits.
\newblock In \emph{International Conference on Machine Learning}, pp.\  3589--3597. PMLR, 2017.

\bibitem[Xie et~al.(2019)Xie, Ma, and Wang]{xie2019towards}
Xie, T., Ma, Y., and Wang, Y.-X.
\newblock Towards optimal off-policy evaluation for reinforcement learning with marginalized importance sampling.
\newblock In \emph{Advances in Neural Information Processing Systems}, pp.\  9665--9675, 2019.

\end{thebibliography}
\bibliographystyle{icml2023}

\clearpage
\onecolumn
\appendix

\section{Related Work} \label{app:related}

\paragraph{Off-Policy Evaluation:}
Off-policy evaluation of counterfactual policies has recently garnered significant interest in both contextual bandits~\citep{dudik2014doubly,farajtabar2018more,kallus2020optimal,kiyohara2022doubly,kiyohara2023off,metelli2021subgaussian,saito2021counterfactual,su2020doubly,su2019cab,wang2017optimal,kiyohara2024off} and reinforcement learning (RL)~\citep{jiang2016doubly,kallus2020double,liu2018breaking,liu2020understanding,thomas2016data,xie2019towards,kiyohara2024towards}. The literature encompasses three main approaches. The first approach, named the Direct Method (DM), is defined as:
\begin{align*}
    \dm := \meanN \mE_{\pi(a|x_i)} [\hat{q} (x_i,a)] = \meanN \sumA \pi(a\,|\,x_i) \hat{q} (x_i,a), 
\end{align*}
where $\hat{q}(x,a)$ estimates $q(x,a)$ based on logged bandit data. This approach exhibits lower variance than IPS and has been utilized to address violations of full support~\citep{sachdeva2020off}, where IPS can be severely biased. However, DM is often vulnerable to reward function misspecification. This issue is problematic, as the extent of misspecification cannot be easily detected and evaluated for real-world data due to non-linearity or partial observability of the environment~\citep{farajtabar2018more,sachdeva2020off,voloshin2019empirical}. The second approach is IPS, which estimates the value of $\pi$ by re-weighting the observed rewards as
\begin{align*}
    \ips := \meanN \frac{\pi(a_i\,|\,x_i)}{\pi_0 (a_i\,|\,x_i)} r_i = \meanN w(x_i,a_i) r_i,
\end{align*}
where $w(x,a) := \pi(a\,|\,x)/\pi_0(a\,|\,x)$ is called the \textit{(vanilla) importance weight}. Under some identification assumptions such as no interference, full support, and no unobserved confounders, IPS provides unbiased and consistent estimation of the value of new policies. However, this approach has a critical drawback: it can suffer from high bias and variance in the presence of numerous actions. First, high bias can occur when the logging policy fails to provide full support (Condition~\ref{assumption:full_support}), which is likely in larger action spaces~\citep{sachdeva2020off,saito2022off}. Furthermore, its variance can be particularly excessive for large action spaces, as the importance weights are prone to taking extremely large values. It is possible to apply weight clipping~\cite{su2020doubly,su2019cab,swaminathan2015counterfactual} and self-normalization~\citep{swaminathan2015self} to somewhat alleviate the variance issue, however, they introduce additional bias in return. DR, which is given as follows, is a third approach that can be considered a hybrid of the previous two approaches, achieving lower bias than DM and lower variance than IPS~\citep{dudik2014doubly,farajtabar2018more}.
\begin{align*}
    \dr := \meanN \left\{ w(x_i,a_i)  (r_i-\hat{q}(x_i,a_i) ) + \mE_{\pi(a|x_i)} [\hat{q} (x_i,a)] \right\}
\end{align*}
Several recent studies have extended DR to further improve its finite sample accuracy~\citep{su2020doubly,wang2017optimal,metelli2021subgaussian} or its robustness to model misspecification~\citep{farajtabar2018more,kallus2020optimal}. Although there is a number of extensions of DR in both bandits (as described above) and RL~\citep{jiang2016doubly,kallus2020double,thomas2016data}, these variants of DR still face the critical variance issue in large action spaces due to the same reasons as IPS~\cite{saito2022off,saito2023off}.

To address the fundamental issues of typical OPE estimators for large action spaces, \citep{saito2022off} proposed a new framework and estimator called Marginalized IPS (MIPS). This approach leverages auxiliary information about the actions, called action embeddings or action features, which are available in many potential applications of OPE such as recommender systems, and provide useful structure in the action space. More specifically, MIPS is defined as:
\begin{align*}
    \mips := \meanN \frac{\pi(e_i\,|\,x_i)}{\pi_0(e_i\,|\,x_i)} r_i = \meanN w(x_i,e_i) r_i,
\end{align*}
where the logged dataset $\calD= \{(x_i,a_i, e_i, r_i)\}_{i=1}^n$ now contains action embeddings for each data point\footnote{$(x,a,e,r) \sim p(x)\pi_0(a\,|\,x)p(e\,|\,x,a)p(r\,|\,x,a,e)$ where $p(e\,|\,x,a)$ is an action embedding distribution.} and $w(x,e) := \frac{\pi(e\,|\,x)}{\pi_0(e\,|\,x)} = \frac{\sum_a p(e\,|\,x,a) \pi(a\,|\,x)}{\sum_a p(e\,|\,x,a) \pi_0(a\,|\,x)}$ is the \textit{marginal importance weight}. This weight is defined with respect to the marginal distributions of the action embeddings induced by the target and logging policies. This enhanced weighting scheme results in significantly lower variance compared to IPS and DR in larger action spaces, while maintaining unbiasedness under the no direct effect assumption. This assumption necessitates that the given action embeddings be informative enough to mediate every causal effect of the actions on the rewards (i.e., $a \perp r\,|\,x,e$). A similar condition regarding the causal structure has been utilized to address the deficient support problem in OPE~\citep{felicioni2022off,lee2022off,peng2023offline,sachdeva2023off} and to conduct causal inference of long-term outcomes through short-term proxies~\citep{athey2020combining,athey2019surrogate,chen2021semiparametric}. However, MIPS may still exhibit high variance, similarly to IPS, when the provided action embeddings are high-dimensional and fine-grained. Additionally, it may generate substantial bias if the no direct effect condition is violated and action embeddings fail to explain much of the causal effects of the actions. This bias issue is particularly expected when performing action feature selection on high-dimensional action embeddings to reduce variance~\cite{su2020adaptive,udagawa2023policy}.

To circumvent the bias-variance dilemma of MIPS, \citep{saito2023off} proposed a more general formulation and a refined estimator. Specifically, instead of relying on the often demanding no direct effect condition, \citep{saito2023off} introduced the conjunct effect model (CEM) of the reward function. The CEM is a useful decomposition of the expected reward function into what is called the cluster effect and residual effect. Building on the CEM, we can employ model-free estimation utilizing cluster importance weights to estimate the cluster effect without bias, and apply model-based estimation using the pairwise regression procedure to estimate the residual effect with low variance as
\begin{align*}
    \offcem := \meanN \left\{ w(x_i,c_{a_i}) (r_i - \hat{f}(x_i,a_i)) + \mE_{\pi(a|x_i)}[\hat{f}(x_i,a)] \right\},
\end{align*}
where $w(x,c) := \frac{\pi(c \,|\,x)}{\pi_0(c \,|\,x)} = \frac{\sumA \mathbb{I} \{ c_a = c \} \pi(a \,|\,x) }{\sumA \mathbb{I} \{ c_a = c \} \pi_0(a \,|\,x) }$ is referred to as the \textit{cluster importance weight}. The first term of OffCEM estimates the cluster effect through cluster importance weighting, while the second term addresses the residual effect using the regression model $\hat{f}$, which is ideally learned via a two-step procedure similar to POTEC. As a result, the OffCEM estimator is likely to achieve significantly lower variance than IPS, DR, and MIPS in scenarios with many actions or high-dimensional action embeddings, while often reducing the bias of MIPS since OffCEM does not ignore the residual effect. Our OPL algorithm is inspired by this CEM formulation, and suggests training two distinct policies via policy-based (model-free) and regression-based (model-based) approaches, respectively.

\paragraph{Off-Policy Learning:}
The contextual bandit framework has emerged as a favored approach for online learning and decision-making under uncertainty~\citep{lattimore2020bandit}, spurring the development of numerous efficient algorithms for navigating (potentially vast or infinite) action spaces~\cite{agrawal2013thompson,li2010contextual}. There is also a growing demand for an offline strategy that refines decision-making without the need for risky and time-consuming active exploration. Consequently, the creation of effective off-policy learning methods in the contextual bandit framework has attracted considerable attention recently~\cite{sachdeva2020off,saito2021counterfactual}. Many real-world interactive systems can capitalize on logged interaction data to learn and enhance a policy offline, enabling safe improvements to the current system's performance~\cite{joachims2018deep,london2019bayesian,sachdeva2020off,saito2021counterfactual,swaminathan2015batch,swaminathan2015counterfactual}.

As already described in Section~\ref{sec:formulation}, there are two main families of approaches in OPL: regression-based and policy-based methods. The regression-based approach relies on a reduction to supervised learning, where a regression estimate is trained to predict the rewards from the logged data~\cite{jeunen2021pessimistic,sachdeva2020off}. To derive a policy, the action with the highest predicted reward is chosen deterministically, or a distribution can be formed based on the estimated rewards as well. A drawback of this straightforward approach is the bias that arises from the misspecification of the regression model. On the other hand, the policy-based approach aims to update the parameterized policy $\pi_{\theta}$ by performing gradient ascent iterations of the form: $\theta_{t+1} \leftarrow \theta_t + \nablat V(\pi_{\theta})$ at each step $t$ during policy learning. Since the true policy gradient $\nablat V (\pi_{\theta}) (= \mE_{p(x)\pi_{\theta}(a|x)}[q(x,a)\nablat\log\pi_{\theta}(a\,|\,x)]$) is unknown, it must be estimated from the logged data using OPE techniques, such as IPS (Eq.~\eqref{eq:is-pg}) and DR (Eq.~\eqref{eq:dr-pg}). However, these estimators necessitate the assumption that the logging policy has full support for every policy in the policy space. This assumption is frequently violated in large action spaces, leading to significant bias in gradient estimation. Moreover, existing policy gradient estimators heavily rely on the vanilla importance weight with respect to the original (potentially large) action space, resulting in critical variance issues and inefficient off-policy learning. One possible approach to address the variance issue in OPE is to apply conservative or imitation regularization~\cite{jeunen2021pessimistic,liang2022local,ma2019imitation,swaminathan2015counterfactual}, which penalize policies that diverge from the logging policy. However, in large action spaces, these regularization techniques often yield a policy that is too close to the logging policy. To tackle the challenges associated with OPE in large action spaces, \citep{lopez2021learning} recently proposed the following selective IPS (sIPS) estimator to estimate the policy gradient.
\begin{align}
   \sipsPG  
   := \meanN  \frac{\pi_{\theta}(a_i\,|\,x_i,a_i\in\Phi(x_i))}{\pi_0(a_i\,|\,x_i)} r_i  \nablat\log \pi_{\theta} (a_i\,|\,x_i), \label{eq:sis-pg}
\end{align}
where $\Phi(x) := \{a\in\calA\,|\,q(x,a) > 0\}$ is the set of relevant actions called the action selector. The idea is to reduce the variance in importance weighting by focusing only on relevant actions assuming that there are many irrelevant actions that have (almost) zero expected rewards in real applications. However, we argue that the variance reduction effect of sIPS is often limited, as it still relies on the logging policy in the denominator. Furthermore, a reliable method for identifying the action selector has not yet been provided.

To address the limitations of existing approaches, we utilize the CEM from~\citet{saito2023off} and proposed the POTEC algorithm, which is the first OPL framework to unify regression-based and policy-based approaches. This algorithm trains two separate policies using regression-based and policy-based approaches, respectively.\footnote{Note that DR in Eq.~\eqref{eq:dr-pg} should be classified as a policy-based approach since its aim is to accurately estimate the true policy gradient, even though it employs a regression-based reward function estimator to achieve variance reduction from IPS.} In particular, our POTEC algorithm is expected to outperform typical policy- and regression-based approaches in large action spaces. First, we utilize cluster importance weighting when training the 1st-stage policy and a regression-based approach when training the 2nd-stage policy, which should yield significantly lower variance compared to existing policy-based methods that apply importance weighting over the original action space. Furthermore, our algorithm is likely to be more robust to reward function misspecification than the regression-based approach, as it relies on a provably unbiased policy gradient in the 1st-stage and aims to estimate only the relative value difference in the 2nd-stage. This is arguably a simpler task compared to the absolute value regression of the conventional regression-based approach.

Note that in the context of reinforcement learning (RL), there are some related ideas and methods to improve sample-efficiency in large action spaces. For example, \citet{chandak2019learning} propose a method to learn action representation to improve sample-efficiency of on-policy RL. However, the focus of \citet{chandak2019learning} is not offline policy learning, and thus its proposed method is not considered as a baseline in our paper. In addition, the supervised representation learning procedure of this paper uses the structure specific to RL (i.e., state transition), so it cannot be applied to our contextual bandit setup. In addition, \citet{gu2022learning} study offline RL in large action spaces and propose a method to learn latent representation in the action space. However, the proposed method of \citet{gu2022learning} leverages the data-distributional metric to learn action embeddings to deal with large action spaces in offline RL, but the metric is based on the MDP structure, and how to apply the method to the offline contextual bandit problem was not discussed and it is non-trivial.

\begin{table*}[h]
\begin{minipage}{\textwidth}
    \caption{Examples of locally correct regression models} \label{tab:example_regression_models}
    \vspace{0.1in}
  \begin{minipage}[t]{.32\textwidth}
    \centering
    \scalebox{0.8}{
    \begin{tabular}{c|cc|cc}
        \toprule
        $a$ & $a_0$ & $a_1$ & $a_2$ & $a_3$ \\ \midrule
        $\phi(x_0,a)$ & \multicolumn{2}{c|}{0} & \multicolumn{2}{c}{1}  \\
        $q(x_0,a)$ & 4 & 1 & 3 & 2  \\
        $\hat{f}_1(x_0,a)$ & 3 & 0 & 1 & 0  \\ \midrule
        $\Delta(x_0,a,b) $ & \multicolumn{2}{c|}{3} & \multicolumn{2}{c}{1}  \\ 
        \bottomrule
    \end{tabular}}
  \end{minipage}
  \hfill
  \begin{minipage}[t]{.32\textwidth}
    \centering
    \scalebox{0.8}{
    \begin{tabular}{c|cc|cc}
        \toprule
        $a$ & $a_0$ & $a_1$ & $a_2$ & $a_3$ \\ \midrule
        $\phi(x_0,a)$ & \multicolumn{2}{c|}{0} & \multicolumn{2}{c}{1}  \\
        $q(x_0,a)$ & 4 & 1 & 3 & 2  \\
        $\hat{f}_2(x_0,a)$ & 50 & 47 & -30 & -31  \\ \midrule
        $\Delta(x_0,a,b) $ & \multicolumn{2}{c|}{3} & \multicolumn{2}{c}{1}  \\ 
        \bottomrule
    \end{tabular}}
  \end{minipage}
  \hfill
  \begin{minipage}[t]{.32\textwidth}
    \centering
    \scalebox{0.8}{
    \begin{tabular}{c|cc|cc}
        \toprule
        $a$ & $a_0$ & $a_1$ & $a_2$ & $a_3$ \\ \midrule
        $\phi(x_0,a)$ & \multicolumn{2}{c|}{0} & \multicolumn{2}{c}{1}  \\
        $q(x_0,a)$ & 4 & 1 & 3 & 2  \\
        $\hat{f}_3(x_0,a)$ & 4 & 1 & 3 & 2  \\ \midrule
        $\Delta(x_0,a,b) $ & \multicolumn{2}{c|}{3} & \multicolumn{2}{c}{1}  \\ 
        \bottomrule
    \end{tabular}}
  \end{minipage}
\end{minipage}
\end{table*}
\section{Examples: Locally Correct Regression Models} \label{app:local_correctness}

This section provides some examples of regression model $\hat{f}$ that satisfies Condition~\ref{assumption:local_correctness} (local correctness). Suppose that there is only a single context $\calX = \{x_0\}$ and four actions $\calA = \{a_0, a_1, a_2, a_3\}$. The expected reward function $q(x,a)$ and clustering function $\phi(x,a)$ are given as follows.
\begin{align*}
    q(x_0,a_0) = 4, \; q(x_0,a_1) = 1, \; q(x_0,a_2) = 3, \; q(x_0,a_3) = 2,\\
    \phi(x_0,a_0) = 0, \; \phi(x_0,a_1) = 0, \; \phi(x_0,a_2) = 1, \; \phi(x_0,a_3) = 1.
\end{align*}
Then, Table~\ref{tab:example_regression_models} provides three locally correct regression models ($\hat{f}_1$ to $\hat{f}_3$). 
More specifically, these example models succeed in preserving the relative value difference of the actions within each action cluster ($c=0$ for $ a_0,a_1$ and $c=1$ for $a_2,a_3$). In fact, we can see that $\Delta_q(x_0,a_0,a_1) = \Delta_{\hat{f}_1}(x_0,a_0,a_1) = \Delta_{\hat{f}_2}(x_0,a_0,a_1) = \Delta_{\hat{f}_3}(x_0,a_0,a_1) = 3$ and $\Delta_q(x_0,a_2,a_3) = \Delta_{\hat{f}_1}(x_0,a_2,a_3) = \Delta_{\hat{f}_2}(x_0,a_2,a_3) = \Delta_{\hat{f}_3}(x_0,a_2,a_3) =  1$ where $\phi(x_0,a_0)=\phi(x_0,a_1)$ and $\phi(x_0,a_2)=\phi(x_0,a_3)$.

\section{Generalization of Our Framework and POTEC Algorithm} \label{app:generalization}
In this section, we describe the generalization of our framework and algorithm to the situation under the presence of some predefined action representation $\phi: \calX \times \calA \rightarrow \calE \subseteq \mathbb{R}^d$, which is often available in practice and can be used to better parameterize the policy. Under the presence of such action representations, we can first generalize the CEM as follows.
\begin{align}
    q(x,a) = \underbrace{g(x,c(x,\Phi(x,a)))}_{\textit{cluster effect}} + \underbrace{h(x,\Phi(x,a))}_{\textit{residual effect}}, \label{eq:general_cem}
\end{align}
where $c: \calX \times \calE \rightarrow \calC$ provides a discretization in the action representation space $\calE$. Note also that the residual effect depends on the representation of the action $\Phi(x,a)$ rather than the atomic actions $a$ as in a simpler version presented in the main text.

Leveraging this general version of the CEM in Eq.~\eqref{eq:general_cem}, we can generalize our POTEC gradient estimator in Eq.~\eqref{eq:potec-pg} in the following two ways.

\paragraph{Implementation Option 1:} This option trains a parameterized distribution over the action representation space $\calE$ as the 1st-stage policy via the following version of the POTEC gradient estimator.
\begin{align}
   \potec  := \meanN \bigg\{ w(x_i,c_i) & (r_i - \hat{f} (x_i,\Phi(x_i,a_i)) ) \nabla_{\theta} \log \pi_{\theta} (\Phi(x_i,a_i)\,|\,x_i) \notag \\
   & + \mE_{e\sim\firstpolicy} [ \hat{f}^{\secondpolicy} (x_i,c) \nabla_{\theta} \log \pi_{\theta} (e\,|\,x_i) ] \bigg\} \label{eq:general-potec-pg1} ,
\end{align}
where $c_i = c(x_i,\Phi(x_i,a_i)), \hat{f}^{\secondpolicy} (x,c) := \mE_{\secondpolicy} [\hat{f}(x,a)]$ and
\begin{align*}
    w(x,c) := \frac{\firstpolicy(c\,|\,x)}{\pi_0^{1st}(c\,|\,x)} = \frac{\int_{e:c(x,e)=c} \firstpolicy(e\,|\,x)}{\int_{e:c(x,e)=c} \pi_0^{1st}(e\,|\,x)}.
\end{align*}
This general version of the POTEC gradient estimator is unbiased under local correctness (i.e, $\Delta_q(x,a,b)=\Delta_{\hat{f}}(x,a,b),\, \forall x,a,b$ such that $c(x,\Phi(x,a))=c(x,\Phi(x,b))$). Since the 1st-stage policy is learned in the action representation space, it can naturally exploit the smoothness in $\calE$.

If we follow this implementation, in the inference time, for an incoming context $x$, we first sample a point in the action representation space $\calE$ from the 1st-stage policy as $e \sim \firstpolicy(\cdot\,|\,x)$, which implies a promising region in $\calE$. Note that, in general, $e \in \calE$ will not match with any already observed action representation $\{\Phi(x_i,a_i)\}_{i=1}^n$. Then, the second-stage $\secondpolicy$, which is constructed from the pairwise regression model $\hat{h}_{\psi}: \calX \times \calE \rightarrow \mathbb{R}$, identifies the best action within the promising region as $$a = \argmax_{a':c(x,\Phi(x,a'))=c(x,e)}\,\hat{h}_{\psi}(x,\Phi(x,a')),$$
where $\{a'\in\calA\,|\,c(x,\Phi(x,a'))=c(x,e)\}$ is the set of actions whose representation lies in the promising region induced by $e \sim \firstpolicy(\cdot\,|\,x)$.

\paragraph{Implementation Option 2:} This option first learns a parameterized distribution over the action space $\calA$ as the 1st-stage policy using $\Phi(x,a)$ as its input via the following version of the POTEC gradient estimator.
\begin{align}
   \potec  := \meanN \bigg\{ w(x_i,c_i) & (r_i - \hat{f} (x_i,\Phi(x_i,a_i))) \nabla_{\theta} \log \pi_{\theta} (a_i\,|\, x_i;\Phi(x_i,a_i)) \notag \\
   & + \mE_{a\sim\firstpolicy} [ \hat{f}^{\secondpolicy} (x_i,c) \nabla_{\theta} \log \pi_{\theta} (a\,|\, x_i;\Phi(x_i,a)) ] \bigg\} \label{eq:general-potec-pg2} ,
\end{align}
where $c_i = c(x_i,\Phi(x_i,a_i)), \hat{f}^{\secondpolicy} (x,c) := \mE_{\secondpolicy} [\hat{f}(x,a)]$ and
\begin{align*}
    w(x,c) := \frac{\firstpolicy(c\,|\,x)}{\pi_0^{1st}(c\,|\,x)} = \frac{\sum_{a:c(x,\Phi(x,a))=c} \firstpolicy(a\,|\,x;\Phi(x,a))}{\sum_{a:c(x,\Phi(x,a))=c} \pi_0^{1st}(a\,|\,x;\Phi(x,a))}.
\end{align*}
This version is also unbiased under local correctness (i.e, $\Delta_q(x,a,b)=\Delta_{\hat{f}}(x,a,b),\, \forall x,a,b$ such that $c(x,\Phi(x,a))=c(x,\Phi(x,b))$). The 1st-stage policy also simply leverages the action representation as its input.\footnote{For example, we can define a parameterized policy as $$\pi_{\theta} (a\,|\, x;\Phi(x,a)) = \frac{\exp(f_{\theta} (x,\Phi(x,a)))}{\sum_{a'\in\calA} \exp(f_{\theta} (x,\Phi(x,a')))}$$ where $f_{\theta}: \calX \times \calE \rightarrow \mathbb{R}$ is some parameterized function having action representation $\Phi(x,a)$ as its input.} 

If we follow this implementation, in the inference time, for an incoming context $x$, we first sample a point in the action space $\calA$ from the 1st-stage policy as $a \sim \firstpolicy(\cdot\,|\,x;\Phi(x,a))$, which merely implies a promising region in $\calE$. Then, the second-stage $\secondpolicy$, which is constructed from the pairwise regression model $\hat{h}_{\psi}: \calX \times \calE \rightarrow \mathbb{R}$, identifies the best action within the promising region as $$a = \argmax_{a':c(x,\Phi(x,a'))=c(x,\Phi(x,a))}\,\hat{h}_{\psi}(x,\Phi(x,a')),$$
where $\{a'\in\calA\,|\,c(x,\Phi(x,a'))=c(x,\Phi(x,a))\}$ is the set of actions whose representation lies in the promising region induced by $a \sim \firstpolicy(\cdot\,|\,x;\Phi(x,a))$.

The empirical comparison of the above two options highly depends on each application. For example, \textbf{Implementation Option 1} may perform better when the action representation space $\calE$ is low-dimensional while it may suffer when $\calE$ is high-dimensional. Therefore, under the presence of some action representation $\Phi(x,a)$, we would encourage the practitioners to identify the best implementations for their particular application in a data-driven fashion, for example, by performing a careful cross-validation.

\subsection{The One-Stage Variant of POTEC} \label{sec:one-stage-variant}
It is worth noting that there exists a one-stage variant of POTEC, as opposed to the two-stage variant, which is our primary proposal. More specifically, the one-stage variant directly trains a parameterized overall policy in the action space, $\pi_{\theta}(a\,|\,x)$, via the POTEC gradient estimator as follows:
\begin{align*}
   \potecone  := \meanN \bigg\{ w(x_i,c_{a_i}) (r_i - \hat{f} (x_i,a_i) ) \score(x_i,a_i) + \mE_{\pi_{\theta}(a|x_i)} [ \hat{f} (x_i,a) \score(x_i,a) ] \bigg\},
\end{align*}
where $\score(x,a) := \nablat\log \pi_{\theta} (a\,|\,x)$. Although the one-stage variant is categorized as a policy-based approach, as it trains the overall policy directly via policy gradient, it still achieves significant variance reduction compared to IPS-PG and DR-PG and remains unbiased under local correctness. However, the one-stage variant could be considered a suboptimal utilization of the local correctness condition since, given a locally correct regression model, we should be able to optimally choose the action within a cluster as in Eq.~\eqref{eq:2nd-stage-policy} and thus do not need to learn the overall policy solely through policy gradient. Nevertheless, the one-stage variant may be valuable in practice, as it do not need to maintain and execute multiple policies. We provide an empirical comparison of the one-stage and two-stage variants of POTEC in Appendix~\ref{app:experiment}.

\section{Omitted Proofs} \label{app:proof}

\subsection{Derivation of Eq.~\eqref{eq:true-1st-stage-pg}} \label{sec:true-1st-stage-pg}
\begin{align*}
    \nabla_{\theta} V\left(\overallpolicy\right) 
    & = \mE_{p(x)}\left[\sumA q(x,a) \nabla_{\theta} \overallpolicy (a\,|\,x)\right] \\
    & = \mE_{p(x)} \left[\sumA q(x,a) \sumC \nabla_{\theta} \firstpolicy(c\,|\,x) \secondpolicy(a\,|\,x, c) \right]\\
    & = \mE_{p(x)} \left[ \sumC \nabla_{\theta} \firstpolicy(c\,|\,x) \sumA q(x,a) \secondpolicy(a\,|\,x, c) \right]\\
    & = \mE_{p(x)} \left[ \sumC \firstpolicy(c\,|\,x) \nabla_{\theta} \log \firstpolicy(c\,|\,x) q^{\secondpolicy}(x, c) \right]\\
    & = \mE_{p(x)\firstpolicy(c\,|\,x)} \left[  q^{\secondpolicy}(x, c) \score(x,c) \right]
\end{align*}
where we use $q^{\secondpolicy}(x, c):= \mE_{\secondpolicy(a|x,c)}[q(x,a)]$ and $\score(x,c):= \nabla_{\theta} \log \firstpolicy(c\,|\,x)$. The above policy gradient suggests increasing the choice probability of a cluster that is promising under the given 2nd-stage policy $\secondpolicy$ where the effectiveness of a cluster under the 2nd-stage policy is quantified by $q^{\secondpolicy}(x, c)$. This implies that the optimal cluster can be different given different 2nd-stage policies. A toy example in Table~\ref{tab:dependence} shows that the value of a cluster can indeed be very different given different 2nd-stage policies. More specifically, the left table shows the case with the optimal 2nd-stage policy that can identify the best action within each cluster. Then, we can see that the optimal cluster is $c=1$, since the maximum expected reward in the actions of this cluster is larger. In contrast, the right table shows the case with uniform 2nd-stage policy. Under such a 2nd-stage policy, the optimal cluster then becomes $c=0$, since the average expected reward of the actions in $c=0$ is larger than that of $c=1$. 

\begin{table*}[t]
\begin{minipage}{\textwidth}
    \caption{Dependence of the cluster value on the 2nd-stage policy ($q^{\secondpolicy}(x, c)$)} \label{tab:dependence}
    \vspace{0.1in}
  \begin{minipage}[t]{.45\textwidth}
    \centering
    \scalebox{1.0}{
    \begin{tabular}{c|cc|cc}
        \toprule
        $a$ & $a_0$ & $a_1$ & $a_2$ & $a_3$ \\ \midrule
        $c(x_0,a)$ & \multicolumn{2}{c|}{0} & \multicolumn{2}{c}{1}  \\
        $q(x_0,a)$ & 4 & 2 & 5 & 0  \\
        $\secondpolicy(a|x,c)$ & 1 & 0 & 1 & 0  \\ \midrule
        $q^{\secondpolicy}(x, c)$ & \multicolumn{2}{c|}{4} & \multicolumn{2}{c}{5}  \\ 
        \bottomrule
    \end{tabular}}
  \end{minipage}
  \hfill
  \begin{minipage}[t]{.45\textwidth}
    \centering
    \scalebox{1.0}{
    \begin{tabular}{c|cc|cc}
        \toprule
        $a$ & $a_0$ & $a_1$ & $a_2$ & $a_3$ \\ \midrule
        $c(x_0,a)$ & \multicolumn{2}{c|}{0} & \multicolumn{2}{c}{1}  \\
        $q(x_0,a)$ & 4 & 2 & 5 & 0  \\
        $\secondpolicy(a|x,c)$ & 0.5 & 0.5 & 0.5 & 0.5  \\ \midrule
        $q^{\secondpolicy}(x, c)$ & \multicolumn{2}{c|}{3} & \multicolumn{2}{c}{2.5}  \\ 
        \bottomrule
    \end{tabular}}
  \end{minipage}
\end{minipage}
\end{table*}

Below we prove the theorems presented in the main text based on the following general version of the POTEC gradient estimator.
\begin{align*}
    \potec  := \meanN \bigg\{ w(x_i,c_i) (r_i - \hat{f} (x_i,a_i) ) \score(x_i,c_{a}) + \mE_{\firstpolicy} [ \hat{f}^{\secondpolicy} (x_i,c) \score(x_i,c_i) ] \bigg\}
\end{align*}
where $c_i \sim p(\cdot\,|\,x_i,a_i)$ is a stochastic and context-dependent clustering. The POTEC gradient estimator defined in Eq.~\eqref{eq:potec-pg} can be considered a special case with a deterministic and context-independent clustering function $c: \calA \rightarrow \calC$.

Note that we use $ w(x,c) = \mE_{\pi(a|x,c)}[w(x,a)]$ and $w(x,a) = \frac{\pi(a\,|\,x)}{\pi_0(a\,|\,x)} = \frac{\pi(a,c\,|\,x)}{\pi_0(a,c\,|\,x)}$ in the following.

\subsection{Proof of Theorem~\ref{thm:bias} and Corollary~\ref{cor:unbiased}} \label{app:bias}
\begin{proof}
To derive the bias of the POTEC gradient estimator, we calculate the difference between its expectation and the true policy gradient given in Eq.~\eqref{eq:true-1st-stage-pg} below.

\begin{align*}
    & \biasPOTEC \\
    & = \mE_{p(x)\pi_0(a|x)p(c|x,a)p(r|x,a)} [w(x,c) (r - \hat{f} (x,a) ) \score(x,c)] + \mE_{p(x)\firstpolicy(c|x)} [ \hat{f}^{\secondpolicy} (x,c) \score(x,c) ] \\
    & \quad - \mE_{p(x)\firstpolicy(c|x)} \left[ q^{\secondpolicy}(x,c) \score(x,c) \right] \\
    & = \mE_{p(x)} \left[ \sumA \pi_0(a\,|\,x) \Delta_{q,\hat{f}} (x,a) \sumC p(c\,|\,x,a)  w(x,c) \score(x,c) \right] + \mE_{p(x)} \left[ \sumC \firstpolicy(c\,|\,x) \hat{f}^{\secondpolicy} (x,c) \score(x,c) \right] \\
    & \quad - \mE_{p(x)} \left[ \sumC \firstpolicy(c\,|\,x) q^{\secondpolicy}(x,c) \score(x,c) \right] \\
    & = \mE_{p(x)} \left[ \sumA \pi_0(a\,|\,x) \Delta_{q,\hat{f}} (x,a) \sumC \frac{\pi_0^{1st}(c\,|\,x)\pi_0^{2nd}(a\,|\,x,c)}{\pi_0(a\,|\,x)} w(x,c) \score(x,c) \right]  \\
    & \quad + \mE_{p(x)} \left[ \sumC \pi_0^{1st}(c\,|\,x) \frac{\firstpolicy(c\,|\,x)}{\pi_0^{1st}(c\,|\,x)} \hat{f}^{\secondpolicy} (x,c) \score(x,c) \right] - \mE_{p(x)} \left[ \sumC \pi_0^{1st}(c\,|\,x) \frac{\firstpolicy(c\,|\,x)}{\pi_0^{1st}(c\,|\,x)} q^{\secondpolicy}(x,c) \score(x,c) \right] \\
    & = \mE_{p(x)\pi_0^{1st}(c|x)} \left[ w(x,c) \score(x,c)  \sumA  \pi_0^{2nd}(a\,|\,x,c) \Delta_{q,\hat{f}} (x,a)\right]  \\
    & \quad + \mE_{p(x)\pi_0^{1st}(c|x)} \left[ w(x,c) \score(x,c) \hat{f}^{\secondpolicy} (x,c) \right] - \mE_{p(x)\pi_0^{1st}(c\,|\,x)} \left[ w(x,c) \score(x,c) q^{\secondpolicy}(x,c) \right] \\
    & = \mE_{p(x)\pi_0^{1st}(c|x)} \left[ w(x,c) \score(x,c)  \sumA  \pi_0^{2nd}(a\,|\,x,c) \Delta_{q,\hat{f}} (x,a)\right]  \\
    & \quad - \mE_{p(x)\pi_0^{1st}(c|x)} \left[ \score(x,c) \sumA \frac{\firstpolicy(c\,|\,x)}{\pi_0^{1st}(c\,|\,x)} \frac{\secondpolicy(a\,|\,x,c)}{\pi_0^{2nd}(a\,|\,x,c)}\pi_0^{2nd}(a\,|\,x,c)  \Delta_{q,\hat{f}} (x,a)  \right] \\
    & = \mE_{p(x)\pi_0^{1st}(c|x)} \left[ \score(x,c) \sumA w(x,a) \pi_0^{2nd}(a\,|\,x,c)  \sum_{b\in\calA}  \pi_0^{2nd}(b\,|\,x,c) \Delta_{q,\hat{f}} (x,b)\right]  \\
    & \quad - \mE_{p(x)\pi_0^{1st}(c|x)} \left[ \score(x,c) \sumA w(x,a) \pi_0^{2nd}(a\,|\,x,c) \Delta_{q,\hat{f}} (x,a)  \right] \\
    & = \mE_{p(x)\pi_0^{1st}(c|x)} \left[ \score(x,c) \sumA w(x,a) \pi_0^{2nd}(a\,|\,x,c) \left( \Big(\sum_{b\in\calA}  \pi_0^{2nd}(b\,|\,x,c) \Delta_{q,\hat{f}} (x,b)\Big) - \Delta_{q,\hat{f}} (x,a) \right)\right]  \\
\end{align*}
where $\Delta_{q,\hat{f}} (x,a):= q(x,a) - \hat{f}(x,a)$. By applying Lemma B.1 of \citep{saito2022off} to the last line (setting $f(a) = w(·, a), g(a) = \pi_0^{2nd}(a\,|\,·, ·), h(a) = \Delta(·, a)$), we obtain the following expression of the bias.
\begin{align*}
    \mE_{p(x)\pi_0^{1st}(c|x)} \left[ \score(x,c) \sum_{a<b} \pi_0^{2nd}(a\,|\,x,c) \pi_0^{2nd}(b\,|\,x,c) \left( \Delta_{q,\hat{f}} (x,a) - \Delta_{q,\hat{f}} (x,b) \right) \left( w(x,b) - w(x,a) \right)  \right]
\end{align*}
In particular, in the simpler case of deterministic and context-independent clustering as in the main text, we can simplify the expression of the bias as below.
\begin{align*}
    & \mE_{p(x)\pi_0^{1st}(c|x)} \left[\sum_{a<b:c_a=c_b=c} \pi_0^{2nd}(a\,|\,x,c) \pi_0^{2nd}(b\,|\,x,c) \left( \Delta_{q,\hat{f}} (x,a) - \Delta_{q,\hat{f}} (x,b) \right) \left( w(x,b) - w(x,a) \right) \score(x,c) \right]\\
    & = \mE_{p(x)\pi_0^{1st}(c|x)} \left[\sum_{a<b:c_a=c_b=c} \pi_0^{2nd}(a\,|\,x,c) \pi_0^{2nd}(b\,|\,x,c) \left( \Delta_{q} (x,a,b) - \Delta_{\hat{f}} (x,a,b) \right) \left( w(x,b) - w(x,a) \right) \score(x,c) \right]
\end{align*}
where, we used $ \pi_0^{2nd}(a\,|\,x,c) = \frac{\pi_0(a\,|\,x) \mathbb{I}\{c_a=c\}}{\pi_0^{1st}(c\,|\,x)} $ and $ \Delta_{q,\hat{f}} (x,a) - \Delta_{q,\hat{f}} (x,b)  \Rightarrow \Delta_{q} (x,a,b) - \Delta_{\hat{f}} (x,a,b)$.
\end{proof}

\subsection{Proof of Proposition~\ref{prop:variance}} \label{app:variance}
\begin{proof}
We apply the law of total variance several times to obtain the variance of the $j$-th element of the POTEC gradient estimator for a particular parameter $\theta \in \mathbb{R}^d$ in the following.
\begin{align*}
    & \mV_{p(x)\pi_0(a|x)p(c|x,a)p(r|x,a)} \left[ w(x,c) (r - \hat{f} (x,a) ) \score^{(j)}(x,c) + \mE_{\firstpolicy(c'|x)} [ \hat{f}^{\secondpolicy} (x,c') \score^{(j)}(x,c') ] \right] \\
    & = \mE_{p(x)\pi_0(a|x)p(c|x,a)} \left[ \mV_{p(r|x,a)} \left[w(x,c) (r - \hat{f} (x,a) ) \score^{(j)}(x,c) + \mE_{\firstpolicy(c'|x)} [ \hat{f}^{\secondpolicy} (x,c') \score^{(j)}(x,c') ] \right] \right] \\
    & \quad + \mV_{p(x)\pi_0(a|x)p(c|x,a)} \left[ \mE_{p(r|x,a)} \left[w(x,c) (r - \hat{f} (x,a) ) \score^{(j)}(x,c) + \mE_{\firstpolicy(c'|x)} [ \hat{f}^{\secondpolicy} (x,c') \score^{(j)}(x,c') ] \right] \right]\\
    & = \mE_{p(x)\pi_0(a|x)p(c|x,a)} \left[ (w(x,c) \score^{(j)}(x,c))^2 \sigma^2(x,a) \right] \\
    & \quad + \mV_{p(x)\pi_0(a|x)p(c|x,a)} \left[ w(x,c) \Delta_{q,\hat{f}}(x,a) \score^{(j)}(x,c) + \mE_{\firstpolicy(c'|x)} [ \hat{f}^{\secondpolicy} (x,c') \score^{(j)}(x,c') ] \right] \\
    & = \mE_{p(x)\pi_0(a|x)p(c|x,a)} \left[ (w(x,c) \score^{(j)}(x,c))^2 \sigma^2(x,a) \right] \\
    & \quad + \mE_{p(x)\pi_0(a|x)} \left[ \mV_{p(c|x,a)} \left[ w(x,c) \Delta_{q,\hat{f}}(x,a) \score^{(j)}(x,c) + \mE_{\firstpolicy(c'|x)} [ \hat{f}^{\secondpolicy} (x,c') \score^{(j)}(x,c') ] \right] \right] \\
    & \quad + \mV_{p(x)\pi_0(a|x)} \left[ \mE_{p(c|x,a)} \left[ w(x,c) \Delta_{q,\hat{f}}(x,a) \score^{(j)}(x,c) + \mE_{\firstpolicy(c'|x)} [ \hat{f}^{\secondpolicy} (x,c') \score^{(j)}(x,c') ] \right] \right] \\
    & = \mE_{p(x)\pi_0(a|x)p(c|x,a)} \left[ (w(x,c) \score^{(j)}(x,c))^2 \sigma^2(x,a) \right]  
    + \mE_{p(x)\pi_0(a|x)} \left[ \mV_{p(c|x,a)} \left[ w(x,c) \Delta_{q,\hat{f}}(x,a) \score^{(j)}(x,c)  ] \right] \right] \\
    & \quad + \mV_{p(x)\pi_0(a|x)} \left[ \mE_{p(c|x,a)} \left[ w(x,c) \Delta_{q,\hat{f}}(x,a) \score^{(j)}(x,c) \right] + \mE_{\firstpolicy(c'|x)} [ \hat{f}^{\secondpolicy} (x,c) \score^{(j)}(x,c') ] \right] \\
    & = \mE_{p(x)\pi_0(a|x)p(c|x,a)} \left[ (w(x,c) \score^{(j)}(x,c))^2 \sigma^2(x,a) \right]  
    + \mE_{p(x)\pi_0(a|x)} \left[ \mV_{p(c|x,a)} \left[ w(x,c) \Delta_{q,\hat{f}}(x,a) \score^{(j)}(x,c)  ] \right] \right] \\
    & \quad + \mE_{p(x)} \left[ \mV_{\pi_0(a|x)} \left[ \mE_{p(c|x,a)} \left[ w(x,c) \Delta_{q,\hat{f}}(x,a) \score^{(j)}(x,c) \right] + \mE_{\firstpolicy(c'|x)} [ \hat{f}^{\secondpolicy} (x,c') \score^{(j)}(x,c') ] \right] \right] \\
    & \quad + \mV_{p(x)} \left[ \mE_{\pi_0(a|x)} \left[ \mE_{p(c|x,a)} \left[ w(x,c) \Delta_{q,\hat{f}}(x,a) \score^{(j)}(x,c) \right] + \mE_{\firstpolicy(c'|x)} [ \hat{f}^{\secondpolicy} (x,c') \score^{(j)}(x,c') ] \right] \right] \\
    & = \mE_{p(x)\pi_0(a|x)p(c|x,a)} \left[ (w(x,c) \score^{(j)}(x,c))^2 \sigma^2(x,a) \right]  
    + \mE_{p(x)\pi_0(a|x)} \left[ \mV_{p(c|x,a)} \left[ w(x,c) \Delta_{q,\hat{f}}(x,a) \score^{(j)}(x,c)  ] \right] \right] \\
    & \quad + \mE_{p(x)} \left[ \mV_{\pi_0(a|x)} \left[ \mE_{p(c|x,a)} \left[ w(x,c) \Delta_{q,\hat{f}}(x,a) \score^{(j)}(x,c) \right] \right] \right] 
    + \mV_{p(x)} \left[ \mE_{\firstpolicy(c|x)} \left[q^{\secondpolicy} (x,c) \score^{(j)}(x,c)\right] \right] ,
\end{align*}
where we rely on local correctness in the last line to use 
\begin{align*}
    \mE_{\pi_0(a|x)} \left[ \mE_{p(c|x,a)} \left[ w(x,c) \Delta_{q,\hat{f}}(x,a) \score^{(j)}(x,c) \right] + \mE_{\firstpolicy(c'|x)} [ \hat{f}^{\secondpolicy} (x,c') \score^{(j)}(x,c') ] \right]  =  \mE_{\firstpolicy(c|x)} \left[q^{\secondpolicy} (x,c) \score^{(j)}(x,c)\right].
\end{align*} 
In particular, in the case of deterministic and context-independent clustering, the variance can be simplified as follows.
\begin{align*}
    & \mV_{p(x)\pi_0(a|x)p(r|x,a)} \left[ w(x,c_a) (r - \hat{f} (x,a) ) \score^{(j)}(x,c_a) + \mE_{\firstpolicy(c'|x)} [ \hat{f}^{\secondpolicy} (x,c) \score^{(j)}(x,c') ] \right] \\
    & = \mE_{p(x)\pi_0(a|x)} \left[ (w(x,c_a) \score^{(j)}(x,c_a))^2 \sigma^2(x,a) \right] \\
    & \quad + \mE_{p(x)} \left[ \mV_{\pi_0(a|x)} \left[  w(x,c_a) \Delta_{q,\hat{f}}(x,a) \score^{(j)}(x,c_a) \right] \right] + \mV_{p(x)} \left[ \mE_{\firstpolicy(c|x)} \left[ q^{\secondpolicy} (x,c) \score^{(j)}(x,c) \right] \right].
\end{align*}
\end{proof} 

\begin{table*}[t]
\centering
\caption{Hyperparameter search spaces used in the experiments. $\lambda$ is the hyperparameter for weight decay. $\eta$ is the learning rate. $B$ is the batch size.}
\vspace{1mm}
\scalebox{0.85}{
\begin{tabular}{cl|c|c|c|c}
\toprule
\multicolumn{1}{l}{Datasets} & Methods & $\lambda$ & $\eta$ & $B$ & $|\Phi(x)|$ in Eq.\eqref{eq:sis-pg} \\ \hline \hline
\multirow{3}{*}{Synthetic} 
 & IPS-PG & $\{10^{-2},10^{-4},10^{-6}\}$ & $\{10^{-3},5\times 10^{-4},10^{-4}\}$ & $\{64,128,256\}$ & $\{0.1|\calA|,0.5|\calA|,|\calA|\}$ \\ 
 & DR-PG & $\{10^{-2},10^{-4},10^{-6}\}$ & $\{10^{-3},5\times 10^{-4},10^{-4}\}$ & $\{64,128,256\}$ & $\{0.1|\calA|,,0.5|\calA|,|\calA|\}$ \\ 
 & POCEM &  $10^{-4}$ & $5\times 10^{-4}$ & $128$ & - \\ \hline
\multirow{3}{*}{Real-World} 
 & IPS-PG & $[10^{-4}, 10^{-2}]$ & $[10^{-4}, 10^{-2}]$ & $1,024$ & $|\calA|$ \\ 
 & DR-PG & $[10^{-4}, 10^{-2}]$ & $[10^{-4}, 10^{-2}]$ & $1,024$ & $|\calA|$ \\ 
 & POCEM &  $10^{-4}$ & $10^{-3}$ & $1,024$ & - \\ 
 \bottomrule
\end{tabular}}
\label{tab:hyperparam}
\end{table*}
\begin{figure*}[h]
\centering
\begin{minipage}{\hsize}
    \begin{center}
        \includegraphics[clip, width=14cm]{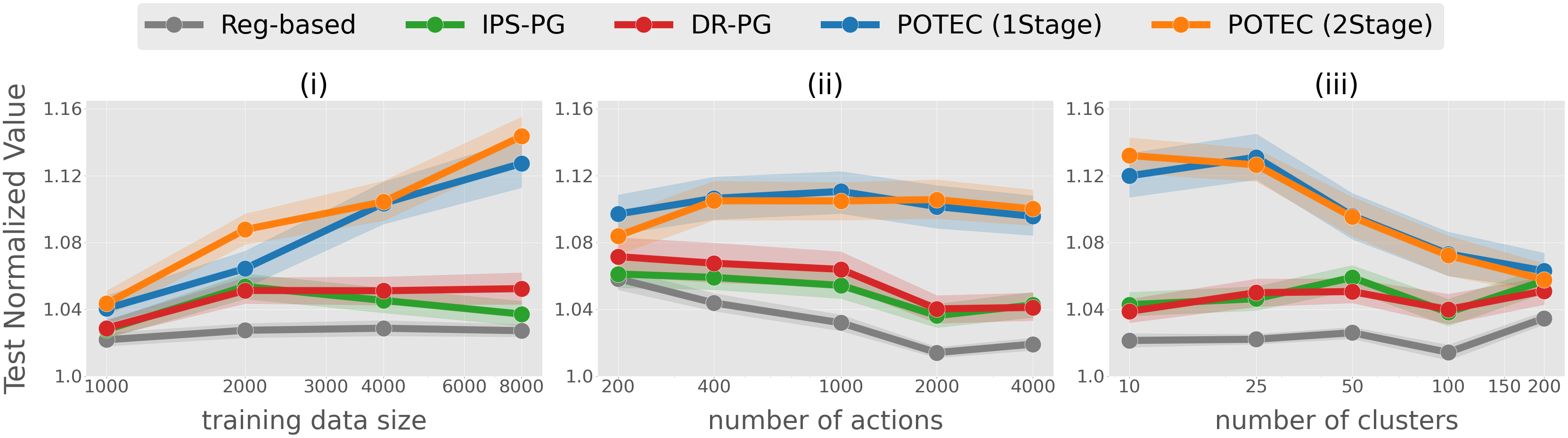}
    \end{center}
    \vspace{-2mm}
    \caption{Comparing the test policy value of the OPL methods with varying \textbf{(i)} training data sizes, \textbf{(ii)} numbers of actions, and \textbf{(iii)} numbers of clusters in the synthetic experiment.}
    \label{fig:main-app}
\end{minipage}
\\ \vspace{5mm}
\begin{minipage}{\hsize}
    \begin{center}
        \includegraphics[clip, width=14cm]{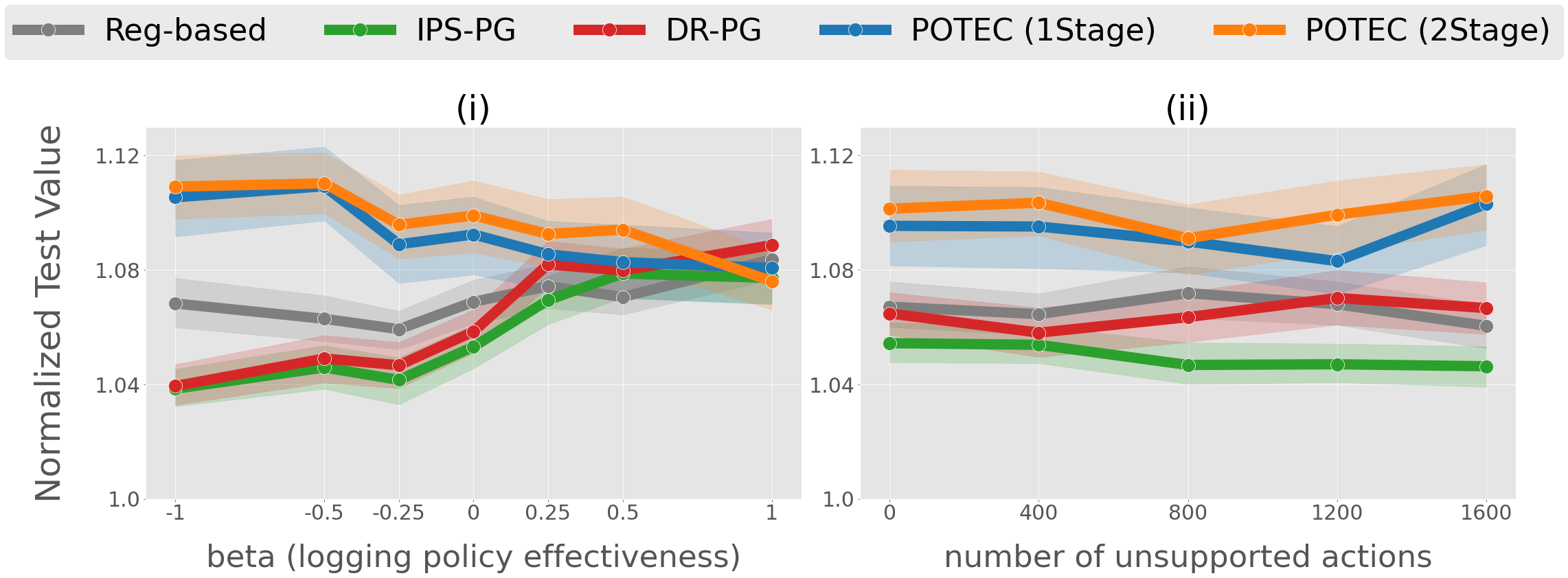}
    \end{center}
    \vspace{-2mm}
    \caption{Comparing the test policy value of the OPL methods with varying \textbf{(i)} logging policies, \textbf{(ii)} numbers of unsupported actions, and \textbf{(iii)} cluster noise ratios in the synthetic experiment.}
    \label{fig:main2-app}
\end{minipage}
\\ \vspace{5mm}
\begin{minipage}{\hsize}
    \begin{center}
        \includegraphics[clip, width=14cm]{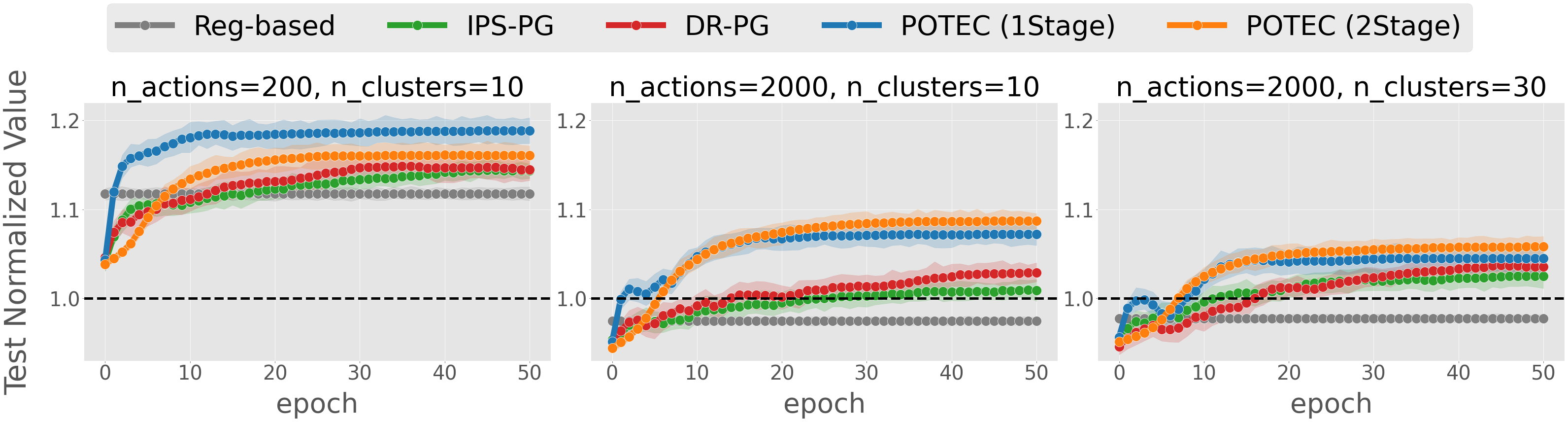}
    \end{center}
    \vspace{-2mm}
    \caption{Comparing the learning curve of the OPL methods when \textbf{(i)} $|\calA|=200,|\calC|=10$, \textbf{(ii)} $|\calA|=2,000,|\calC|=10$, and \textbf{(iii)} $|\calA|=2,000,|\calC|=30$ in the synthetic experiment.}
    \label{fig:learning-curve-app}
\end{minipage}
\\ \vspace{5mm}
\begin{minipage}{\hsize}
    \begin{center}
        \includegraphics[clip, width=14cm]{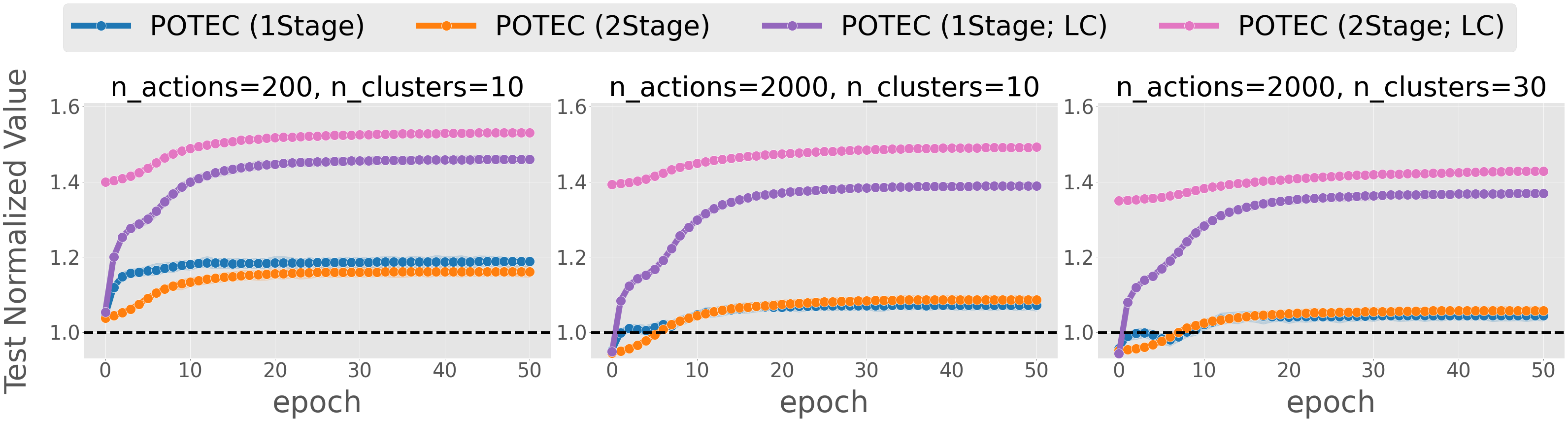}
    \end{center}
    \vspace{-2mm}
    \caption{Comparing the learning curve of the one-stage and two-stage POCEM w/ or w/o a locally correct regression model when \textbf{(i)} $|\calA|=200,|\calC|=10$, \textbf{(ii)} $|\calA|=2,000,|\calC|=10$, and \textbf{(iii)} $|\calA|=2,000,|\calC|=30$ in the synthetic experiment. ``LC'' stands for \textbf{L}ocally \textbf{C}orrect.}
    \label{fig:learning-curve-lc-app}
\end{minipage}
\vskip 0.1in
\raggedright
\fontsize{10pt}{10pt}\selectfont \textit{Note}:
We set $n=4,000$, $|\calA|=2,000$, and $|\calC|=30$ as default experiment parameters.
The results are averaged over 100 different sets of synthetic logged data replicated with different random seeds.
The shaded regions in the plots represent the 95\% confidence intervals of the policy value estimated with bootstrap.
\end{figure*}

\begin{figure*}[h]
\centering
\begin{minipage}{\hsize}
    \begin{center}
        \includegraphics[clip, width=14cm]{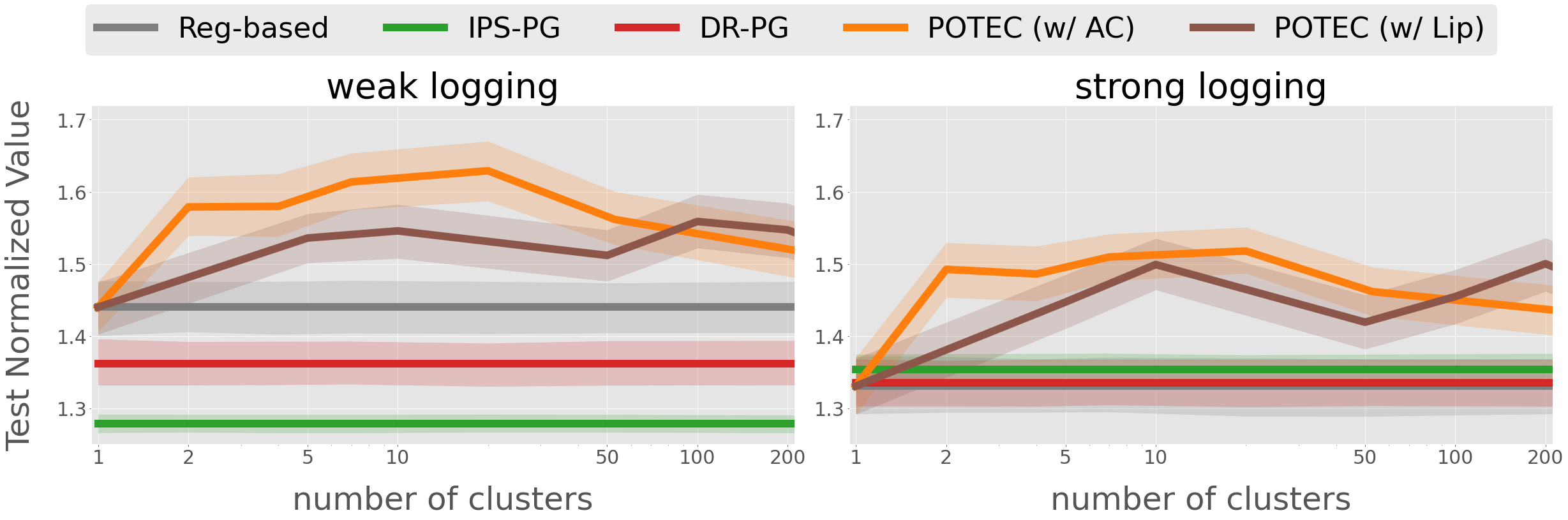}
    \end{center}
    \vspace{-2mm}
    \caption{Comparing the test policy value of the OPL methods (normalized by $V(\pi_0)$) on the Eurlex-4K dataset with weak and strong logging policies, respectively.}
    \label{fig:real-eurlex}
\end{minipage}
\\ \vspace{15mm}
\begin{minipage}{\hsize}
    \begin{center}
        \includegraphics[clip, width=14cm]{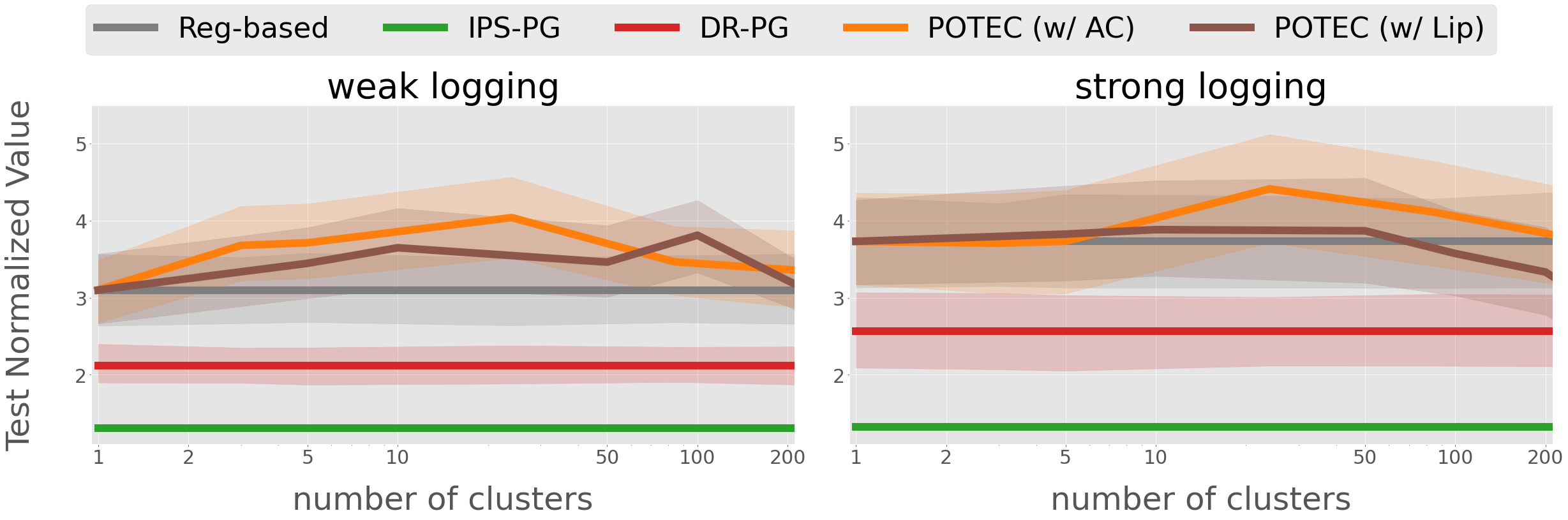}
    \end{center}
    \vspace{-2mm}
    \caption{Comparing the test policy value of the OPL methods (normalized by $V(\pi_0)$) on the Wiki10-31K dataset with weak and strong logging policies, respectively.}
    \label{fig:real-wiki}
\end{minipage}
\vskip 0.1in
\raggedright
\fontsize{10pt}{10pt}\selectfont \textit{Note}:
The results are averaged over 5 different sets of synthetic logged data replicated with different random seeds.
The shaded regions in the plots represent the 95\% confidence intervals of the policy value estimated with bootstrap.
\end{figure*}
\section{Additional Experiment Setups and Results} \label{app:experiment}

\subsection{Synthetic Experiment}
\paragraph{Detailed Setup.} This section describes how we define the synthetic reward function and perform hyperparameter tuning in detail. 
Recall that, in the synthetic experiment, we synthesized the expected reward function as
\begin{align}
    q(x,a) = g(x,c_a ) + h_{c_a}(x,a),  \label{eq:synthetic_reward}
\end{align}
where we use the following functions as $g(\cdot,\cdot)$ (cluster effect) and $h(\cdot,\cdot,\cdot)$ (residual effect), respectively. 
\begin{align*}
    g(x,c_a ) 
    &= g_{base}(x,c_a ) + u_1 \mathbb{I}\{ (\sum_{d=1}^3 x_d) < 1.5 \} \\
    & \quad + u_2 \mathbb{I}\{ (\sum_{d=3}^8 x_d) < -0.5 \} + u_3  \mathbb{I}\{ (\sum_{d=2}^3 x_d) > 3.0 \} + u_4 \mathbb{I}\{ (\sum_{d=5}^{10} x_d) < 1.0 \}, \\
    h_{c_a}(x,a) &= x^{\top} M_{c_a} \text{one\_hot}_{a} + \theta_{x,{c_a}}^{\top} x +  \theta_{a,{c_a}}^{\top} \text{one\_hot}_{a},
\end{align*}
 where $x_d$ is the $d$-th dimension of the context vector $x$. We use \textbf{obp.dataset.polynomial\_reward\_function} from OpenBanditPipeline\footnote{https://github.com/st-tech/zr-obp} as $g_{base}(\cdot,\cdot)$ and $u_1,\ldots,u_4$ are sampled from a uniform distribution with range $[-3,3]$. $M_{c_a}$, $\theta_{x,{c_a}}$, and $\theta_{a,{c_a}}$ are parameter matrices or vectors sampled from a uniform distribution with range $[-1,1]$ separately for each given action cluster $c_a$. 

We synthesized the logging policy $\pi_0$ as
\begin{align}
    \pi_0(a\,|\,x) =  \frac{\exp( \beta \cdot q(x,a) + \mu(x,a))}{ \sum_{a' \in \calA} \exp( \beta \cdot q(x,a') + \mu(x,a)) } , \label{eq:synthetic_logging}
\end{align}
where $\beta$ is a parameter that controls the optimality of the logging policy, and we use $\beta=0$ as default. We use \textbf{obp.dataset.polynomial\_behavior\_policy\_function} from OpenBanditPipeline as $\mu(\cdot,\cdot)$.

To summarize, we first sample a context and define the expected reward $q(x,a)$ as in Eq.~\eqref{eq:synthetic_reward}. We then sample discrete action $a$ from $\pi_0$ based on Eq.~\eqref{eq:synthetic_logging} where action $a$ is associated with a cluster $c_a$. The reward is then sampled from a normal distribution with mean $q(x,a)$. Iterating this procedure $n$ times generates logged data $\calD$ with $n$ independent copies of $(x,a,c_a,r)$.

We tuned the weight decay hyperparameter, learning rate, batch size, and the number of irrelevant actions for variance reduction for the baseline methods (i.e., IPS-PG and DR-PG) using the test policy value, while we use a fixed set of hyperparameters for POTEC as shown in Table~\ref{tab:hyperparam}, giving an unfair advantage to the baselines. For all methods, we used Adam~\citep{kingma2014adam} as an optimizer and used neural networks with 3 hidden layers to parameterize the policy.

Note that the experiments were conducted on MacBook Pro (Apple M2 Max, 96 GB). 

\paragraph{Additional Synthetic Results.}
Figures~\ref{fig:main-app} to~\ref{fig:learning-curve-lc-app} report additional results in the synthetic experiment. Figure~\ref{fig:main-app} compares the test policy value of the OPL methods with varying \textbf{(i)} training data sizes, \textbf{(ii)} numbers of actions, and \textbf{(iii)} numbers of (true) clusters as in the main text, but we additionally compare the one-stage variant of POTEC from Section~\ref{sec:one-stage-variant} with the same regression model as used for the two-stage variant. We can see that the one-stage and two-stage variants of POTEC perform very similarly with a learned regression model, and they both substantially outperform the baseline methods in a range of situations. Figure~\ref{fig:main2-app} reports the results with varying \textbf{(i)} logging policies (a larger $\beta$ means a more effective logging policy, see Eq.~\eqref{eq:synthetic_logging} for the definition of the logging policy), \textbf{(ii)} numbers of unsupported actions ($|\{a \in \calA\,|\,\pi_0 (a\,|\,\cdot) = 0\}|$), and \textbf{(iii)} cluster noise ratios. We can see from the figure that the one-stage and two-stage variants of POTEC perform similarly here as well, and they work much better than the baselines for a range of logging policies and under the violation of full support. POTEC is also still superior to the baseline methods even when 30\% of the true cluster membership is perturbed, demonstrating its robustness to inaccurate action clustering (though it is important to obtain accurate clustering for a more effective OPL). Figure~\ref{fig:learning-curve-app} shows the learning curve of the OPL methods when \textbf{(i)} $|\calA|=200,|\calC|=10$, \textbf{(ii)} $|\calA|=2,000,|\calC|=10$, and \textbf{(iii)} $|\calA|=2,000,|\calC|=30$, where we can see that POTEC stably improves its value throughout the learning process due to its low variance, while IPS-PG and DR-PG have much larger confidence intervals, indicating their unstable learning due to excessive variance in gradient estimation. Figure~\ref{fig:learning-curve-lc-app} compares the one-stage and two-stage variants of POTEC with or without a locally correct (LC) regression model. We can see that the two variants of POTEC perform similarly when combined with a learned regression model, as observed in other results, but the two-stage variant of POTEC performs significantly better than the one-stage variant since the two-stage POTEC optimally utilizes the local correctness condition.

\begin{table}[t]
\caption{Dataset Statistics} \label{tab:data_stats}
\vspace{2mm}
\centering
\scalebox{1}{
\begin{tabular}{c|ccc}
\toprule
\textbf{Dataset} & $n_{train}$ & $n_{test}$ & $|\calA|$ \\\midrule \midrule
EUR-Lex 4K & 15,449 & 3,865 & 3,956 \\
Wiki10-31K & 14,146 & 6,616 & 30,938 \\
\bottomrule
\end{tabular}
}
\vspace{-2mm}
\end{table}
\subsection{Real-World Experiment}

\paragraph{Setup.} Following previous studies~\citep{dudik2014doubly,saito2021evaluating,su2020doubly,wang2017optimal}, we transform the extreme classification datasets to contextual bandit feedback data with many actions. In a classification dataset $\{(x_i,a_i)\}_{i=1}^{n}$, we have some feature vector $x_i \in \calX$ and ground-truth label $a_i \in \calA$, which will be considered an action.

We consider stochastic continuous rewards where we define the expected reward function as follows.
\begin{align}
    \label{eq:real_reward} 
    q(x,a)=
    \left\{\begin{array}{ll}1-\eta_a & \text {if $a$ is a positive label} \\ \eta_a & \text {otherwise}\end{array}\right.
\end{align}
where $\eta_a$ is a noise parameter sampled separately for each action $a$ from a uniform distribution with range $[0,0.1]$. After defining the expected reward function, we sample the reward from a normal distribution as $r \sim \mathcal{N}(q(x,a),\sigma^2)$ with standard deviation $\sigma = 0.05$ for each data.

We define the logging policy $\pi_0$ by applying the softmax function to an estimated reward function $\tilde{q}(x,a)$ as
\begin{align}
    \pi_0(a\,|\,x) =  \frac{\exp( \beta \cdot \tilde{q}(x,a))}{ \sum_{a' \in \calA} \exp( \beta \cdot \tilde{q}(x,a')) },
    \label{eq:real_logging}
\end{align}
where we use $\beta=10$ for both datasets. We obtain $\tilde{q}(x,a)$ by learning a matrix factorization model where we use the test data recorded in the original datasets for obtaining a logging policy while we use the training data for performing OPL to make them independent.

\paragraph{Results.} 
Figures~\ref{fig:real-eurlex} and~\ref{fig:real-wiki} report the test policy value (normalized by $V(\pi_0)$) of the OPL methods with varying numbers of clusters on Eurlex-4K and Wiki10-31K, using two types of logging policies. For these experiments, we trained a "weak logging" policy with (two times) fewer samples than the "strong logging" policy. We optimized the hyperparameters of POTEC and the baselines based on the ground-truth policy value in the validation set, and the effectiveness of the OPL methods is evaluated on the test set. It should be noted that the baseline methods do not depend on action clusters, which results in flat lines in the figures.

The figures demonstrate that POTEC, with both clustering methods (Lipschitz regularization; Lip and Agglomerative clustering; AC, as detailed in the main text), typically outperforms all baseline methods across a range of numbers of clusters. The regression-based method performs competitively with POTEC only for a strong logging policy on the Wiki10-31K dataset, but we can see, in all other scenarios, POTEC typically performs the best. We also compared the one-stage and two-stage variants of POTEC on the real-world datasets, but we did not find a significant difference between them for both types of clustering.

\end{document}